\PassOptionsToPackage{table}{xcolor}
\documentclass{ifm-tr}

\usepackage{amsmath,amsfonts,bm}

\def\eqref#1{equation~\ref{#1}}
\def\1{\bm{1}}

\DeclareMathAlphabet{\mathsfit}{\encodingdefault}{\sfdefault}{m}{sl}
\SetMathAlphabet{\mathsfit}{bold}{\encodingdefault}{\sfdefault}{bx}{n}

\renewcommand{\cite}{\citep}
\usepackage{amsmath,amsfonts,amssymb,mathtools,bm}
\usepackage{amsthm}
\usepackage{thmtools}
\usepackage{algorithm}
\usepackage{algpseudocode}
\usepackage{enumitem}
\usepackage{siunitx}
\usepackage{tikz}
\usetikzlibrary{arrows.meta,fit,backgrounds,positioning}
\usepackage[utf8]{inputenc}
\usepackage{url}
\usepackage{xurl}
\usepackage{booktabs}
\usepackage{nicefrac}
\usepackage{graphicx}
\usepackage{tabularx}
\usepackage{xspace}
\usepackage{multirow}
\usepackage{makecell}
\usepackage{wrapfig}
\usepackage{lipsum}
\ifdefined\XeTeXrevision
  \providecommand{\faLightbulb}{}
\else
  \usepackage{fontawesome5}
\fi
\usepackage{xcolor}
\usepackage{listings}
\newcommand\rurl[1]{\href{https://#1}{\nolinkurl{#1}}}
\newcommand\grurl[1]{\href{https://github.com/#1}{\nolinkurl{#1}}}
\newcommand\hrurl[1]{\href{https://huggingface.co/#1}{\nolinkurl{#1}}}
\newcommand\wrurl[1]{\href{https://wandb.ai/#1}{\nolinkurl{#1}}}

\definecolor{colbg}{HTML}{F2F2F2}
\definecolor{dimrule}{HTML}{CCCCCC}
\definecolor{cellHighlight}{HTML}{dbefff}

\definecolor{blue-primary}{HTML}{0081FB}
\definecolor{blue-light}{HTML}{D4ECFF}
\newtcolorbox{keytakeaways}[2][]{
    enhanced,
    colback=blue-light,
    colframe=blue-primary,
    colbacktitle=blue-primary,
    coltitle=white,
    fonttitle=\bfseries\large\color{white},
    title={\faLightbulb\, #2},
    after=\par\vspace{0.5em},
    arc=2.5pt,
    boxrule=0.6pt,
    left=9pt, right=9pt, top=8pt, bottom=8pt,
    drop fuzzy shadow,
    #1
}

\title{\bf Training-Time Partial Visual Evidence in World-Action Models:\\ An Empirical Study}

\author[1,2]{Haoran Ding}
\author[1,2]{Wen Yang}
\author[1,2]{Tianyu Liu}
\author[1]{Dong An}
\author[2]{Yoshihiko Nakamura}
\author[1]{Guangyi Liu}
\author[2]{Anqing Duan}

\affiliation[1]{Institute of Foundation Models}
\affiliation[2]{Mohamed bin Zayed University of Artificial Intelligence (MBZUAI)}

\date{
    December 1, 2026\\[4pt]
    \faGlobe\ \textbf{Project Page:}\ 
\url{https://mrdinghr.github.io/see-less-act-better/}}

\reportnumber{2026-12-01}

\abstract{
World-action models (WAMs) jointly learn action
prediction and future visual dynamics yet remain
vulnerable to visual distribution shifts.
We study how training-time visual corruption
affects robustness across representation level,
within-view design, and cross-view allocation,
holding supervision, architecture, and test
observations unchanged.
Across 50 RoboTwin and 18 RoboCasa tasks,
we identify three findings.
First, nominally aligned masking does not imply
equivalent robustness: pixel-space corruption
achieves higher aggregate success than the tested
attention-based latent-token scheme on both benchmarks.
Second, spatial organization matters beyond
texture realism: structured diverse occlusion
with shared-color replacement (SDO-SC)
substantially outperforms simpler masking on
RoboTwin but performs comparably to textured SDO
on both benchmarks.
Shared-color replacement also outperforms
independent coloring in aggregate.
Third, broader view coverage does not replace
stronger per-view occlusion: at equal summed
nominal budgets, weaker distributed masking
underperforms stronger single-view corruption
on RoboTwin; full-strength all-view SDO
achieves the highest aggregate success on both
benchmarks.

Across RoboTwin configurations, lower
background-induced action sensitivity tracks
higher OOD success.
All-view SDO raises RoboTwin background-OOD
success from 7.6\% to 58.1\% and mean real-robot
success across four visual shifts from
15.3\% to 68.0\% over 1,500 trials.
These findings guide training-time visual
corruption design for robust WAMs.
}

\begin{document}
\maketitle

% \renewcommand{\thefootnote}{\fnsymbol{footnote}}
% % \footnotetext[1]{Correspondence to: \texttt{\{first.author, second.author\}@mbzuai.ac.ae}}
% \renewcommand{\thefootnote}{\arabic{footnote}}

% ----------------------------------------------------------------------
%  Artifact links (optional)
% ----------------------------------------------------------------------

% \clearpage
% \setcounter{tocdepth}{2}
% \tableofcontents
% \newpage

% ===============================================================%

% ===============================================================%

% ===============================================================%

\section{Introduction}
\label{sec:introduction}

World-action models (WAMs) couple action prediction
with visual world modeling~\cite{ye2026world,yuan2026fast,LiL-RSS-26,bi2026motus}.
Recent models use future-frame prediction as a training signal while shortening or skipping it at test time~\cite{yuan2026fast,zhao2026faster}.
Yet WAMs remain vulnerable to visual distribution shifts. One remedy is to remove part of the visual evidence during training, which raises a central question: \emph{which properties of this removal matter for out-of-distribution (OOD) robustness?}
% Yet WAMs remain vulnerable to visual distribution
% shifts. This raises a central question:
% \emph{which properties of training-time visual
% evidence removal matter for OOD robustness?}

% \begin{figure}[t]
%     \centering
%     \includegraphics[width=\columnwidth]{
%         img/fig1_compact_bars_12pt_gain_inkteal_preview.png
%     }
%     \caption{
%         \textbf{Training with partial visual evidence
%         improves generalization to unmasked OOD inputs.}
%         On 50 RoboTwin tasks with unseen backgrounds,
%         success rises from $7.6\%$ (Standard) to
%         $27.4\%$ (Random Erasing) and $53.4\%$
%         (single-view SDO). The corrupted conditions
%         share the clean-sample probability and masking-ratio distribution.
%     }
%     \label{fig:teaser}
% \end{figure}

Visual robustness has been explored via domain
randomization~\cite{tobin2017domain}, augmentation~\cite{laskin2020reinforcement,hansen2021generalization,
hansen2021stabilizing}, and masking~\cite{zhong2020random,he2022masked}.
Visual RL overlays blend rather than remove evidence~\cite{hansen2021generalization,hansen2021stabilizing},
while WAM robustness has also been addressed through
architecture~\cite{wang2026st}.
How the \emph{form} of removed visual evidence affects
multi-view WAM robustness remains unclear.
We corrupt training-time visual conditioning only;
the action and future-video targets remain clean, and
the architecture and inference are unchanged.

\begin{wrapfigure}{r}{0.55\textwidth}
    \centering
    \vspace{-10pt}
    \includegraphics[width=\linewidth]{
        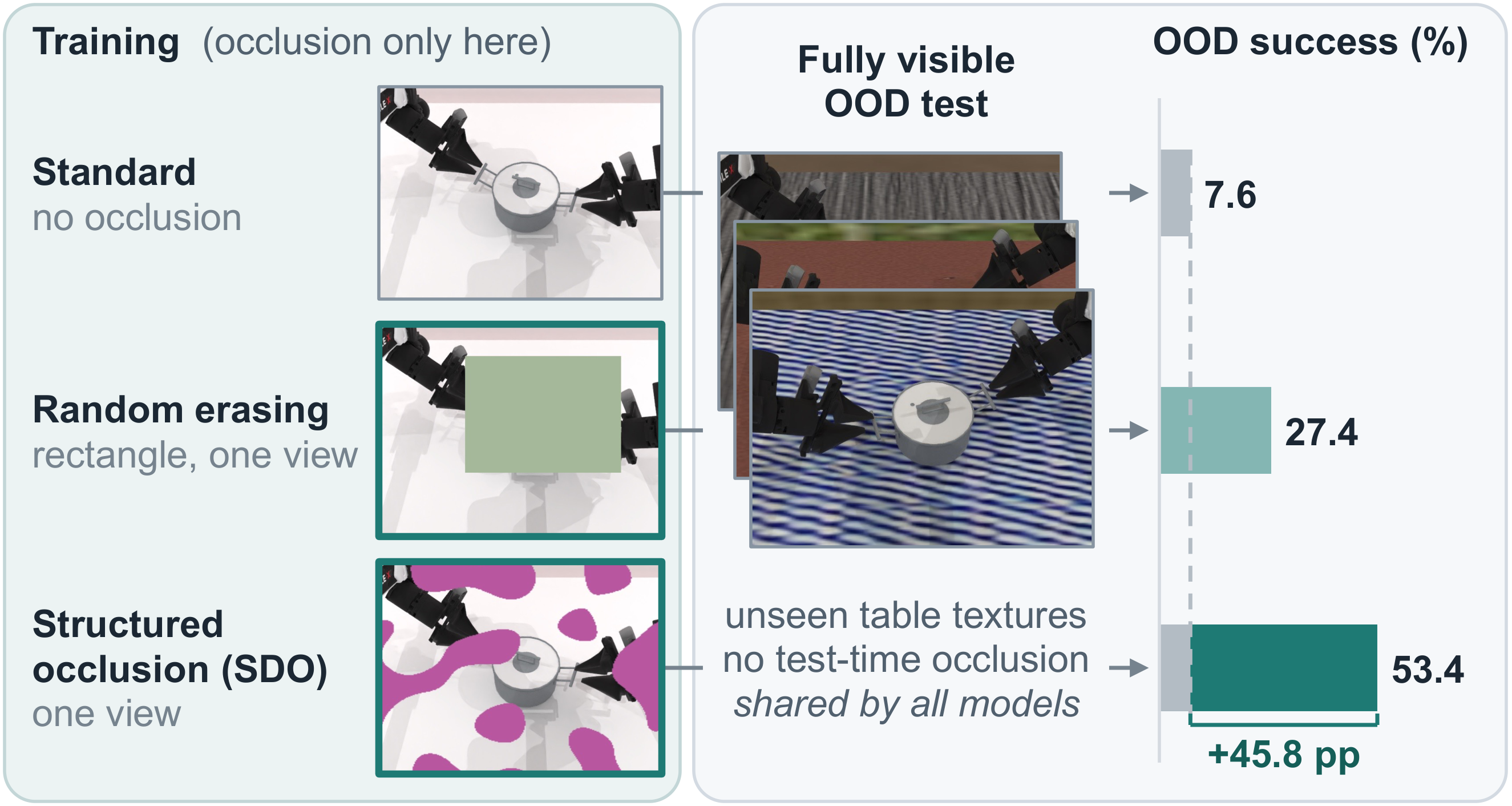
    }
    \caption{
        \textbf{Training with partial visual evidence
        improves generalization to unmasked OOD inputs.}
        On 50 RoboTwin tasks with unseen backgrounds,
        success rises from $7.6\%$ (Standard) to
        $27.4\%$ (Random Erasing) and $53.4\%$
        (single-view SDO). The corrupted conditions
        share the clean-sample probability and masking-ratio distribution.
    }
    \label{fig:teaser}
    \vspace{-8pt}
\end{wrapfigure}

Figure~\ref{fig:teaser} shows that incomplete training
observations can improve robustness to unmasked OOD
inputs. Yet corruption schemes perform differently (27.4\% vs. 53.4\% in Fig.~\ref{fig:teaser}):
\emph{masking alone does not specify the training
intervention.} One might expect nominally aligned
pixel and token masks (same patch footprint and masking ratio) to behave similarly, more
realistic or varied fills to help, and a fixed
masking budget spread across views to be beneficial.
We test these intuitions through three questions:
\textbf{(RQ1)} Does nominally aligned pixel- and
latent-token masking yield comparable robustness?
\textbf{(RQ2)} How do spatial design and replacement
appearance affect robustness?
\textbf{(RQ3)} How should masking be allocated across views, at an equal summed budget and at full strength in every view?

We hold data, supervision, architecture, optimization,
and inference fixed within each comparison.
Across 50 RoboTwin tasks with fixed-background
training and 18 RoboCasa tasks with randomized
backgrounds, we evaluate over 40,000 simulated
episodes and identify three empirical findings.

\textbf{(1) Nominally aligned masking does not
imply equivalent robustness.}
Pixel-space corruption achieves higher aggregate
success than the tested attention-based latent-token
scheme on both benchmarks.

\textbf{(2) Spatial organization matters beyond
texture realism.}
Structured diverse occlusion with shared-color
replacement (SDO-SC) substantially outperforms
simpler masking on RoboTwin. It performs comparably to textured SDO on both benchmarks, and shared-color replacement outperforms independent coloring on average.

\textbf{(3) Broader view coverage does not replace
stronger per-view occlusion.}
At equal summed nominal budgets, concentrating
corruption in one view outperforms distributing
weaker masks across views on RoboTwin. Full-strength all-view SDO, which uses a larger summed budget, achieves the highest mean success on both benchmarks.
On RoboCasa, which trains with randomized backgrounds, the advantages of SDO-SC over simpler masking and of single-view over matched-budget masking are both smaller.

% Both single-view advantages are less pronounced on
% RoboCasa, which uses background-randomized training.

All-view SDO raises RoboTwin background-OOD success
from $7.6\%$ to $58.1\%$ and RoboCasa randomized-background success from $53.9\%$ to $73.3\%$. Across RoboTwin configurations, lower background-induced action sensitivity tracks higher OOD success. We also evaluate SDO under two additional WAM training objectives, where it raises background-OOD success by 47–54 pp, and on a physical robot, where mean success over four visual shifts rises from $15.3\%$ to $68.0\%$ in 1,500 trials. SDO adds no inference-time cost.
Together, these results show that WAM robustness depends not only on whether visual evidence is removed but on how the missing evidence is represented, organized, and allocated across views.

\section{Related Work}
\label{sec:related_work}

\begin{table}[t]
\centering
\scriptsize
\setlength{\tabcolsep}{2pt}
\caption{
\textbf{Prior work involving partial visual evidence,
organized along our three study axes.}
$^\dagger$Reconstruction objective.
$^\ddagger$Blended overlay rather than removal.
}
\label{tab:positioning}
\begin{tabular}{@{}lllll@{}}
\toprule
Method & Model & Level & Within view & Across views \\
\midrule
Cutout~\cite{devries2017improved}, RE~\cite{zhong2020random}
    & classifier & pixel & rectangle & -- \\
H\&S~\cite{kumar2017hide}, GridMask~\cite{chen2020gridmask}
    & classifier & pixel & grid & -- \\
MAE~\cite{he2022masked}, SimMIM~\cite{Xie_2022_CVPR}$^\dagger$
    & ViT & pix./latent & rand.\ patches & -- \\
PatchDropout~\cite{liu2023patchdropout}
    & ViT & latent & rand.\ patches & -- \\
SODA~\cite{hansen2021generalization}, SVEA~\cite{hansen2021stabilizing}
    & RL agent & pixel & overlay$^\ddagger$ & -- \\
MWM~\cite{seo2023masked}$^\dagger$
    & world model & feature & rand.\ patches & -- \\
MV-MWM~\cite{seo2023multi}$^\dagger$
    & world model & pixel & rand.\ patches & masked views \\
3D-MVP~\cite{qian20253d}$^\dagger$
    & manip.\ enc. & pixel & rand.\ patches & all views \\
Sensor Dropout~\cite{liu2017learning}
    & policy & input & whole stream & dropped streams \\
M$^3$~\cite{cheng2026robust}
    & VLA & latent & -- & view/modality mask \\
MaskVLA~\cite{jiang2026maskvla}
    & VLA & pixel & -- & main view \\
\midrule
\textbf{Ours} & \textbf{WAM} & pix.\ / latent & 4 families
    & 1 / matched / all \\
\bottomrule
\end{tabular}
\end{table}

\paragraph{World-action models and visual robustness.}
Vision-language-action (VLA) models leverage pretrained
vision-language representations for robot
control~\cite{pmlr-v229-zitkovich23a,pmlr-v270-kim25c,
pmlr-v305-black25a}.
WAMs additionally learn future
visual dynamics~\cite{zhao2025cot,ye2026world,
LiL-RSS-26,bi2026motus,yuan2026fast,ni2026gigaworld},
with some reducing test-time imagination~\cite{zhao2026faster}.
Visual robustness has been explored through
semantic-temporal modeling~\cite{wang2026st},
visual bottlenecks, perturbation training,
run-time interventions, and physical-shift
evaluations~\cite{pmlr-v306-fu26x,xie2026strong,
hancock2025run,liu2025eva}, as well as progressive
visual sub-goal planning~\cite{zhang2026world_}.
We instead study training-time visual corruption
without changing the WAM architecture.

\paragraph{Visual augmentation for policy generalization.}
Domain randomization~\cite{tobin2017domain},
visual RL augmentation~\cite{laskin2020reinforcement,yarats2021image,
yarats2022mastering}, and augmented world
models~\cite{ball2021augmented}
improve policy generalization.
SODA and SVEA use natural-image overlays in
visual RL~\cite{hansen2021generalization,
hansen2021stabilizing}, while generative
augmentation diversifies demonstration
appearances or viewpoints~\cite{Chen-RSS-23,
chen2025semantically,pmlr-v305-pan25a,
pmlr-v305-jang25a}.
Other work addresses test-time
occlusion~\cite{noh2025sample}.
In contrast, we corrupt conditioning observations only during training, without synthesizing target-domain appearances or masking test inputs.

\paragraph{Masking and partial visual evidence.}
Image masking includes Cutout, Hide-and-Seek,
Random Erasing, GridMask, and CutMix~\cite{devries2017improved,kumar2017hide,
zhong2020random,chen2020gridmask,yun2019cutmix},
alongside masked image modeling~\cite{he2022masked,bao2022beit,Xie_2022_CVPR}
and token dropout~\cite{liu2023patchdropout}.
Occlusion-aware training has also been studied
in video action detection~\cite{modi2023occlusions}.
In robot learning, masked world models~\cite{seo2023masked,seo2023multi}
and multi-view representation pretraining~\cite{qian20253d} mainly use masking for
reconstruction. 
Sensor Dropout removes entire
input streams~\cite{liu2017learning}, while
M$^3$ masks view, language, and action-query
tokens during bimanual VLA training
~\cite{cheng2026robust} and MaskVLA masks
main-camera observations during VLA
fine-tuning~\cite{jiang2026maskvla}.
FroFA examines input- versus feature-space
augmentation~\cite{Bar_2024_CVPR}.
% Sensor Dropout removes entire
% input streams~\cite{liu2017learning}, while
% M$^3$ masks view, language, and action-query
% tokens during bimanual VLA training
% ~\cite{cheng2026robust}.

We compare representation
level, within-view corruption design, and
cross-view allocation for WAMs under unchanged
action and future-video supervision
(Tab.~\ref{tab:positioning}).
Following empirical studies of robot perception
and data choices~\cite{Xue_2026_CVPR,Wen_2025_CVPR},
we distill practical guidelines on where, how,
and across which views to remove visual
evidence during WAM training.

\section{Study Design}
\label{sec:study_design}

\begin{figure*}[t]
    \centering
    \includegraphics[width=\textwidth]{
        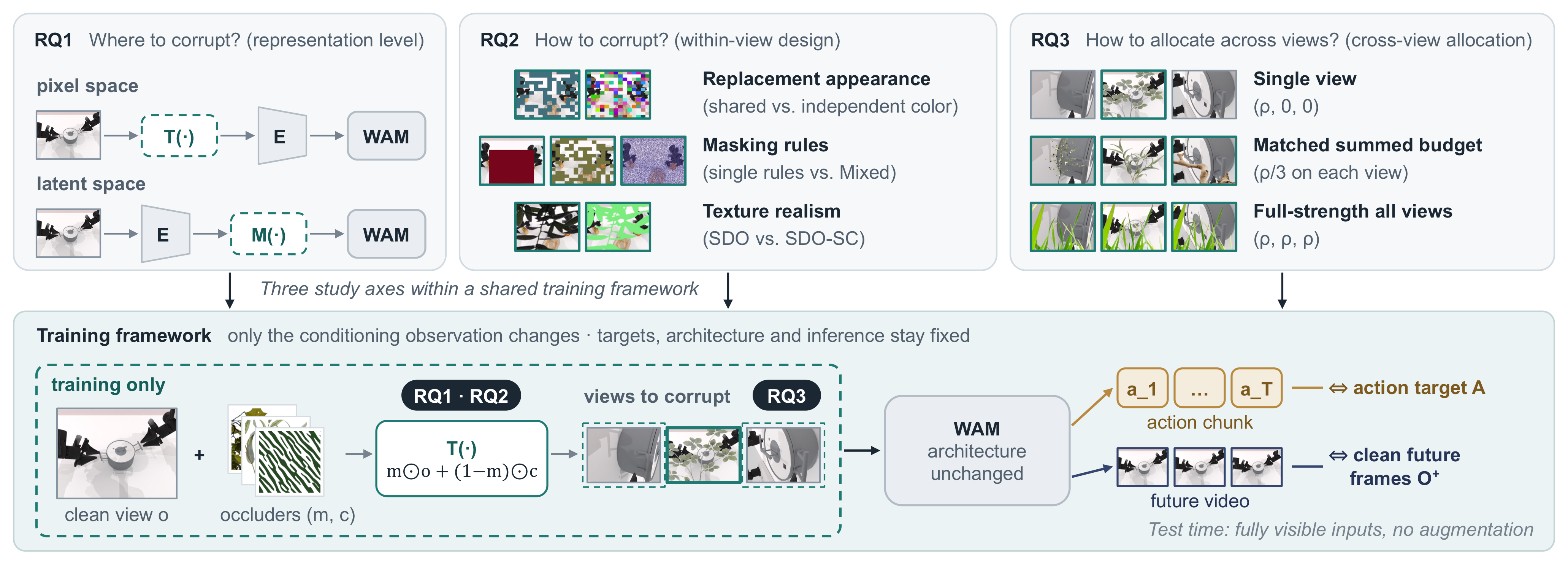
    }
    \caption{
        \textbf{Study design.}
        We examine three dimensions of training-time
        partial visual evidence:
        \textbf{(RQ1)} representation level,
        \textbf{(RQ2)} within-view design, and
        \textbf{(RQ3)} cross-view allocation.
        Only visual conditioning is modified;
        action and future-video targets remain clean.
        Test observations are unmasked.
        The lower panel illustrates pixel-space corruption.
    }
    \label{fig:study_design}
\end{figure*}

\subsection{Training-Time Partial Visual Evidence}
\label{sec:prelim}
\label{sec:partial_evidence}

We consider WAMs that take
multi-view observations $O_t=(I_t^1,\ldots,I_t^V)$
and task instructions to predict an action chunk
$A_t$. Future observations $O_t^+$ provide
additional visual-dynamics supervision.
Our primary model is Fast-WAM~\cite{yuan2026fast},
which combines a Wan2.2 video diffusion
transformer (DiT), a frozen video VAE, a text encoder,
and an action expert through shared attention.
It is trained to predict video and actions
but generates actions directly at inference
without future video prediction.

We modify only visual conditioning during training.
For an image $o\in\mathbb{R}^{H\times W\times C}$,
pixel-space corruption is
\begin{equation}
    T_{\xi}(o)
    =m\odot o+(1-m)\odot c,
    \qquad \xi=(m,c),
    \label{eq:occlusion}
\end{equation}
where $m\in[0,1]^{H\times W}$ is a visibility
mask and $c$ specifies the replacement appearance.
Latent masking instead restricts the attention
visibility of encoded visual tokens.

Let $\widetilde{O}_t$ denote the modified conditioning.
Training minimizes
\begin{equation}
    \begin{aligned}
    \mathcal{L}(\theta)
    =\mathbb{E}_{\tau,\xi}\big[
    &\ell_{\mathrm{act}}
        (\theta;\widetilde{O}_t,A_t)
    &+ \lambda\ell_{\mathrm{world}}
        (\theta;\widetilde{O}_t,O_t^+)
    \big],
    \end{aligned}
    \label{eq:visual_objective}
\end{equation}
where $\tau=(O_t,A_t,O_t^+)$.
Action and future-video targets remain clean,
without an additional reconstruction objective.

During corruption training, each sample is left clean
with probability $p_{\mathrm{clean}}=0.5$.
Otherwise, the masking ratio is
$\rho\sim\mathcal{U}(0,1)$, except for the two RQ1 schemes,
which use $q\sim\mathcal{U}(0,1/3)$ for alignment
with latent masking. Test inputs are unmasked.

\subsection{Study Dimensions}
\label{sec:study_dimensions}

Fig.~\ref{fig:study_design} illustrates the three study dimensions.

\paragraph{RQ1: Representation level.}
\label{sec:representation_design}
We test whether nominally aligned pixel-space
and latent-token masking yield comparable
robustness. Pixel corruption operates before
the frozen VAE; attention-based latent masking
removes selected current-frame visual tokens
from attention.

For alignment, \textbf{MAE-Matched} partitions the composite
three-view observation into $32\times32$-pixel patches,
the nominal footprint of one visual token after
the VAE's $16\times$ spatial downsampling and
the DiT's $2\times2$ spatial patching.
Both schemes use the same selection rule,
clean probability, $q\sim\mathcal{U}(0,1/3)$,
and uniform sampling over the canvas.
Masked pixel patches share one random fill color
per training sample.

\paragraph{RQ2: Within-view corruption design.}
\label{sec:within_view_design}

With pixel-space and single-view corruption
fixed, we examine two properties: the geometry
and spatial distribution of masked regions,
and the appearance used to replace them.

\emph{Occlusion geometry and spatial distribution.}
\textbf{Random Erasing (RE)}~\cite{zhong2020random}
replaces a random rectangle with a solid color.
\textbf{MAE-SC} selects random patches on a
$14\times14$ grid, while \textbf{Bernoulli-SC}
masks pixels independently.
\textbf{Mixed Masking} samples RE, MAE-SC,
or Bernoulli-SC with equal probability.
\textbf{Structured Diverse Occlusion (SDO)}
adapts natural-object cutouts and synthetic
occlusion patterns from the Visual Occluders
Dataset~\cite{coggan2026exposure}.
Both sources are sampled equally.
Natural occluders retain RGB textures, while
synthetic occluders use randomly colored
geometric and procedural patterns.
The dataset contains no manipulation scenes.

\emph{Replacement appearance.}
MAE-SC and Bernoulli-SC use one random RGB
color per sample, whereas their independent
counterparts, \textbf{MAE-style masking}~\cite{he2022masked} and
\textbf{Bernoulli masking}, assign independent
colors per masked patch or pixel, respectively.
Each pair shares the same mask geometry.
\textbf{SDO-SC} preserves SDO's mask sampling,
occlusion shapes, and synthetic branch, but
replaces natural-object textures with a shared
random RGB color.
The clean probability and masking-ratio
distribution remain fixed across these schemes.

\paragraph{RQ3: Cross-view allocation.}
\label{sec:cross_view_design}

Using SDO across $V$ camera views, we compare
three schemes. \emph{Single-view} applies ratio
$\rho$ to one randomly selected view.
\emph{Matched-budget} distributes the same
summed budget, applying $\rho/V$
per view. \emph{All-view} applies $\rho$ to
every view, yielding a total budget $V\rho$.
Single-view vs.\ matched-budget tests concentrating vs.\ spreading occlusion at an equal summed budget, where concentrating means $V\times$ stronger occlusion per view. All-view tests whether full strength in every view helps further at a larger budget.
% it reuses one occluder layout across views,
% whereas matched-budget samples each view independently.
% \paragraph{RQ3: Cross-view allocation.}
% \label{sec:cross_view_design}

% Using SDO across $V$ camera views, we compare
% three schemes. \emph{Single-view} applies ratio
% $\rho$ to one randomly selected view.
% \emph{Matched-budget} distributes the same
% summed budget, applying $\rho/V$
% per view. \emph{All-view} applies $\rho$ to
% every view, yielding a total budget $V\rho$.
% Single-view and matched-budget compare allocation
% at equal summed budgets; all-view
% evaluates full-strength cross-view corruption
% with a larger budget and shared occluder placement.

\subsection{Experimental Protocol}
\label{sec:protocol}

\paragraph{Benchmarks.}
On RoboTwin, we evaluate 50 bimanual tasks from
RoboTwin~\cite{chen2025robotwin}, using one
head and two wrist cameras.
Each task provides 50 demonstrations under
a fixed training background.
Clean evaluation retains the original appearance;
background OOD (BG OOD) uses unseen table and wall textures.
Lighting and clutter shifts are additionally
evaluated for Standard and all-view SDO.
Each policy is evaluated for 30 episodes per task, split across three fixed sets of environment seeds.

On RoboCasa, we evaluate 18 atomic tasks from
RoboCasa~\cite{robocasa365}, whose 9,126
demonstrations already use randomized backgrounds;
Standard here thus serves as a data-level
randomization baseline.
We evaluate 10 episodes per task (180 per condition,
one seed set) under randomized backgrounds,
without a strict seen/unseen background split.

% \textbf{RoboCasa.}
% We evaluate 18 atomic tasks from
% RoboCasa~\cite{robocasa365}, using 9,126
% demonstrations with background-randomized training.
% We evaluate 10 episodes per task under randomized
% backgrounds using one seed set, without a strict
% seen/unseen background split.

\paragraph{Training and reporting.}
Methods share demonstrations, architecture,
objectives, optimization budget, and final-checkpoint
protocol within each comparison.
Each condition uses one training run; $95\%$ confidence intervals (CIs) are obtained by bootstrapping over tasks and reflect task variation.
We further evaluate two WAM training objectives, joint video--action modeling
(Joint) and inverse-dynamics modeling (IDM),
as well as a physical bimanual robot platform.

% We additionally evaluate Joint and IDM
% Fast-WAM configurations and a real bimanual robot.

\paragraph{Background sensitivity.}
We measure action changes when identical simulated
states are re-rendered with different table and
wall textures, holding robot and object states
and camera poses fixed.
For paired observations $O$ and $O^u$, using
identical action-sampling noise $\epsilon$, define
\begin{equation}
    \begin{aligned}
    D_{\mathrm{bg}}(\theta)
    =\mathbb{E}_{O,u,\epsilon}\big[
    d_{\mathrm{nRMS}}\big(
    &a_\theta(O^u;\epsilon),
    &a_\theta(O;\epsilon)
    \big)\big],
    \end{aligned}
    \label{eq:bg_dependency}
\end{equation}
where $d_{\mathrm{nRMS}}$ is the normalized
root-mean-square action-chunk difference.
We examine its association with OOD success
in Sec.~\ref{sec:visual_dependency}.

\section{Experimental Analysis}
\label{sec:experiments}
We answer RQ1–RQ3 on RoboTwin and RoboCasa (Tabs.~\ref{tab:robotwin} and~\ref{tab:robocasa}), then analyze background-induced action sensitivity, test two additional WAM training objectives, and validate the gains on a real bimanual robot. Results under lighting and clutter shifts are in Appendix~\ref{app:additional_shifts}.

% We examine how the representation, within-view
% design, and cross-view allocation of missing
% visual evidence affect WAM robustness
% (RQ1--RQ3), using RoboTwin and RoboCasa
% (Tabs.~\ref{tab:robotwin} and~\ref{tab:robocasa}).
% These comparisons reveal when nominally aligned
% masking differs across representations, how
% spatial design and replacement appearance matter,
% and why cross-view allocation cannot be
% characterized by masking budget alone.
% We further analyze background-induced action
% sensitivity, evaluate robustness across two
% additional WAM configurations, and validate
% the overall robustness gain on a real bimanual robot.
% Additional results under
% lighting and clutter shifts are reported in
% Appendix~\ref{app:additional_shifts}.

\begin{table}[t]
    \centering
    \footnotesize
    \setlength{\tabcolsep}{4pt}
    \renewcommand{\arraystretch}{1.04}
    \caption{
        \textbf{RoboTwin results over 50 tasks.}
        Success rates (\%) are aggregated over
        30 evaluation episodes per task across
        three fixed seed sets.
        BG OOD uses unseen backgrounds and
        unmasked test observations.
        $\Delta$ is the BG OOD gain over
        Standard (pp); SC denotes shared-color
        replacement. MAE-Matched follows the
        token-aligned protocol in
        Sec.~\ref{sec:representation_design}.
    }
    \label{tab:robotwin}
    \begin{tabular*}{0.7\columnwidth}{
        @{\extracolsep{\fill}}lccc@{}}
        \toprule
        \textbf{Condition}
            & \textbf{Clean}
            & \textbf{BG OOD}
            & $\boldsymbol{\Delta}$ \\
        \midrule
        Standard
            & 74.9 & 7.6 & -- \\
        \midrule
        \multicolumn{4}{@{}l@{}}{
            \textbf{RQ1: Representation level}} \\
        \quad Latent
            & 76.3 & 17.0 & +9.4 \\
        \quad MAE-Matched
            & 80.3 & 20.6 & +13.0 \\
        \midrule
        \multicolumn{4}{@{}l@{}}{
            \textbf{RQ2: Within-view design}} \\
        \quad Random Erasing
            & 78.4 & 27.4 & +19.8 \\
        \quad MAE-style
            & 79.2 & 22.2 & +14.6 \\
        \quad Bernoulli
            & 80.3 & 29.5 & +21.9 \\
        \quad MAE-SC
            & 79.4 & 30.1 & +22.5 \\
        \quad Bernoulli-SC
            & \textbf{82.7} & 34.8 & +27.2 \\
        \quad Mixed Masking
            & 79.7 & 36.9 & +29.3 \\
        \quad SDO-SC
            & 80.6 & 50.6 & +43.0 \\
        \quad SDO (single)
            & 79.2 & 53.4 & +45.8 \\
        \midrule
        \multicolumn{4}{@{}l@{}}{
            \textbf{RQ3: Cross-view allocation}} \\
        \quad SDO (matched)
            & 80.9 & 41.2 & +33.6 \\
        \quad SDO (all)
            & 80.2 & \textbf{58.1}
            & \textbf{+50.5} \\
        \bottomrule
    \end{tabular*}
\end{table}

\begin{table}[t]
    \centering
    \footnotesize
    \setlength{\tabcolsep}{5pt}
    \renewcommand{\arraystretch}{1.05}
    \caption{
        \textbf{RoboCasa results over 18 tasks.}
        Success rates (\%) are measured under randomized
        backgrounds (10 episodes per task).
        $\Delta$ is relative to Standard (pp).
        No strict seen/unseen background split is used.
        MAE-Matched follows the token-aligned protocol in
        Sec.~\ref{sec:representation_design}.
    }
    \label{tab:robocasa}
    \begin{tabular*}{0.7\columnwidth}{
        @{\extracolsep{\fill}}lcc@{}}
        \toprule
        Condition & Success & $\Delta$ \\
        \midrule
        Standard & 53.9 & -- \\
        \midrule
        \multicolumn{3}{@{}l}{\emph{RQ1: Representation level}} \\
        \quad Latent
            & 51.1 & $-2.8$ \\
        \quad MAE-Matched
            & 59.4 & $+5.5$ \\
        \midrule
        \multicolumn{3}{@{}l}{\emph{RQ2: Within-view design}} \\
        \quad Random Erasing
            & 58.9 & $+5.0$ \\
        \quad MAE-style
            & 61.1 & $+7.2$ \\
        \quad Bernoulli
            & 63.3 & $+9.4$ \\
        \quad MAE-SC
            & 66.7 & $+12.8$ \\
        \quad Bernoulli-SC
            & 68.9 & $+15.0$ \\
        \quad Mixed Masking
            & 62.2 & $+8.3$ \\
        \quad SDO-SC
            & 70.6 & $+16.7$ \\
        \quad SDO (single)
            & 68.3 & $+14.4$ \\
        \midrule
        \multicolumn{3}{@{}l}{\emph{RQ3: Cross-view allocation}} \\
        \quad SDO (matched)
            & 66.7 & $+12.8$ \\
        \quad SDO (all)
            & \textbf{73.3} & $\mathbf{+19.4}$ \\
        \bottomrule
    \end{tabular*}
\end{table}

\subsection{RQ1: Representation Level}
\label{sec:representation_level}
On RoboTwin, pixel-space MAE-Matched
achieves 20.6\% background-OOD success versus
17.0\% for attention-based latent masking
(Tab.~\ref{tab:robotwin}).
Clean success is also higher (80.3\% vs. 76.3\%).
The same trend holds on RoboCasa:
59.4\% versus 51.1\% (Tab.~\ref{tab:robocasa}).

Per task, MAE-Matched
wins on 27 of 50 RoboTwin tasks, ties on 14, and loses
on 9, a mean gain of 3.6 percentage points (pp; 95\% CI: [1.5, 5.7]).
On RoboCasa, the corresponding counts are
9/4/5, with a larger mean gain of 8.3 pp
but a wider CI of [$-$1.7, 18.9]
(Appendix~\ref{app:statistics}).
Thus, the advantage is more consistent across
RoboTwin tasks, while RoboCasa shows greater
task-level uncertainty.

The two interventions differ beyond nominal mask
statistics. Pixel-space replacement passes through
the frozen VAE, whereas latent masking restricts
encoded visual tokens through attention.
Both use $q\sim\mathcal{U}(0,1/3)$ for alignment.
On RoboTwin, the 3.6 pp representation-level gap
is smaller than the performance ranges observed
within RQ2 (31.2 pp) and RQ3 (16.9 pp).
Thus, corruption design and cross-view allocation
produce substantially greater variation than the
tested representation-level comparison.

\paragraph{Finding 1.}
\textbf{Nominally aligned masking does not imply
equivalent robustness across representations.}
Pixel-space corruption achieves higher mean
success than attention-based latent-token masking
on both benchmarks.

\subsection{RQ2: Within-View Corruption Design}
\label{sec:occlusion_design}
We first compare occlusion designs with simple replacement colors, then test whether richer textures or more diverse colors improve robustness.

% \paragraph{How important is occlusion geometry?}
% Under the same single-view protocol and
% nominal masking-ratio distribution,
% SDO-SC achieves 50.6\% BG OOD success
% on RoboTwin, exceeding Random Erasing
% (27.4\%), MAE-SC (30.1\%), Bernoulli-SC
% (34.8\%), and Mixed Masking (36.9\%)
% (Tab.~\ref{tab:robotwin}).
% Mixed Masking combines three simple rules
% but improves over Bernoulli-SC by only
% 2.1 pp (95\% task-bootstrap CI:
% [-0.7, 5.0]; Appendix~C.3).
% The remaining 13.7 pp gap to SDO-SC
% highlights the importance of structured
% occlusion beyond mixing basic masking rules.

\paragraph{How important is occlusion geometry?}
We compare SDO-SC with simpler masking schemes
under the same single-view protocol and nominal
masking-ratio distribution.
On RoboTwin, SDO-SC achieves 50.6\% BG OOD
success, exceeding Random Erasing
(27.4\%), MAE-SC (30.1\%), Bernoulli-SC (34.8\%),
and Mixed Masking (36.9\%)
(Tab.~\ref{tab:robotwin}).
The gains over Bernoulli-SC and Mixed Masking
are 15.8 and 13.7 pp, respectively.
Mixed Masking combines three simple masking
rules yet exceeds Bernoulli-SC by only 2.1 pp
(95\% task-bootstrap CI: [-0.7, 5.0];
Appendix~\ref{app:statistics}).
Combining simple masking patterns thus does
not close the gap to SDO-SC: occlusion geometry and spatial distribution matter even without realistic textures.

% \paragraph{Does richer replacement appearance help?}
% We next vary replacement appearance while
% preserving the underlying mask geometry.
% First, SDO and SDO-SC share the same occlusion
% shapes, sampling, and synthetic branch,
% differing only in natural-object textures.
% On RoboTwin, textured SDO achieves 53.4\%
% versus 50.6\% for SDO-SC, a difference of
% 2.8 pp (95\% task-bootstrap CI for SDO-SC
% minus SDO: [-6.0, 0.5]).
% On RoboCasa, SDO-SC instead exceeds textured
% SDO (70.6\% vs.\ 68.3\%).
% The effect of adding realistic textures is
% therefore much smaller than the performance
% differences between structured and simpler
% occlusion designs on RoboTwin, with no
% consistent additional gain across benchmarks.

% We further compare shared and independent
% replacement colors under identical mask
% geometry (Appendix Fig.~\ref{fig:app_within_view}).
% MAE-SC and Bernoulli-SC use one random RGB
% color per sample, whereas their independent
% counterparts assign separate colors per masked
% patch or pixel.
% On RoboTwin, shared colors improve BG OOD
% success from 22.2\% to 30.1\% for MAE masking
% and from 29.5\% to 34.8\% for Bernoulli
% (Tab.~\ref{tab:robotwin}).
% The respective gains are 7.9 pp
% (95\% task-bootstrap CI: [5.1, 10.9])
% and 5.3 pp ([2.3, 8.5];
% Appendix~\ref{app:statistics}).
% On RoboCasa, both shared-color variants
% improve aggregate success by 5.6 pp
% (Tab.~\ref{tab:robocasa}).
% Thus, increasing within-image color complexity
% does not improve robustness; simple, coherent
% replacement colors perform better in aggregate.

\paragraph{Does richer replacement appearance help?}
We first compare textured SDO with SDO-SC.
On RoboTwin, textured SDO achieves 53.4\%
BG OOD success versus 50.6\% for SDO-SC
(95\% task-bootstrap CI for SDO-SC
minus SDO: [-6.0, 0.5]).
On RoboCasa, SDO-SC instead achieves 70.6\%
versus 68.3\% for textured SDO.
Prior work found benefits from naturalistic
occluder shapes and textures for object
recognition under test-time occlusion~\cite{coggan2026exposure}.
In our setting, test observations are
unmasked and robustness is evaluated
under background shifts, where natural-object
textures provide no consistent additional
gain across benchmarks.

We next compare shared and independent
replacement colors under identical mask
geometry (Appendix Fig.~\ref{fig:app_within_view}).
On RoboTwin, shared colors improve BG OOD
success from 22.2\% to 30.1\% for MAE
(95\% task-bootstrap CI: [5.1, 10.9] pp)
and from 29.5\% to 34.8\% for Bernoulli
([2.3, 8.5] pp; Appendix~\ref{app:statistics}).
On RoboCasa, both shared-color variants
improve success by 5.6 pp
(Tab.~\ref{tab:robocasa}).
Simple, coherent replacement colors thus
outperform independent coloring on both benchmarks.

\begin{wrapfigure}[17]{r}{0.44\textwidth}
    \centering
    \vspace{-5pt}
    \includegraphics[width=\linewidth]{
        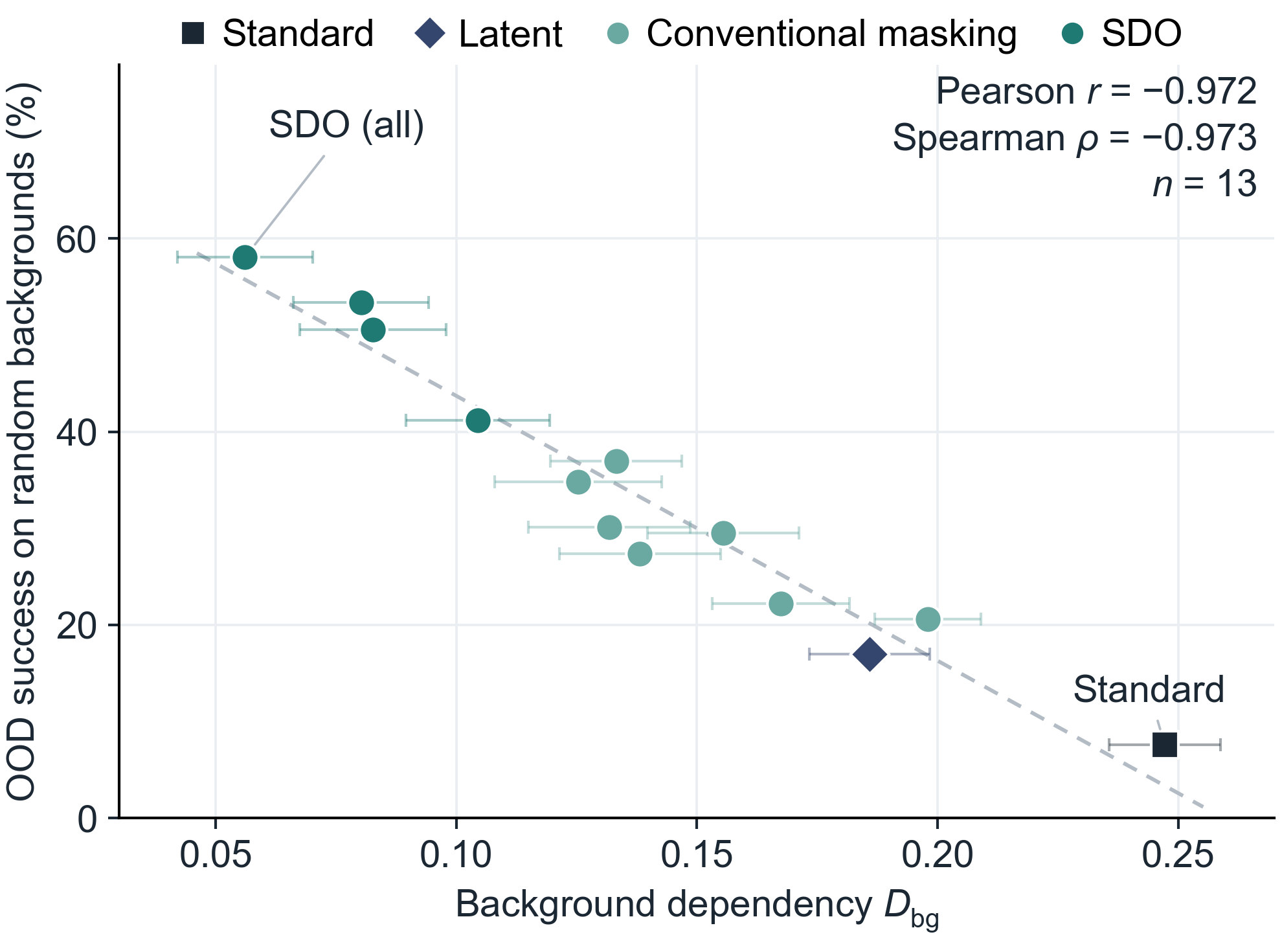
    }
    \caption{
        \textbf{Background-induced action sensitivity
        vs. OOD success.}
        $D_{\mathrm{bg}}$ is measured on 10 randomly
        selected RoboTwin tasks; success is aggregated
        over all 50 tasks.
        Horizontal error bars indicate the standard
        error of $D_{\mathrm{bg}}$; the dashed line
        shows a linear fit across 13 configurations.
    }
    \label{fig:dependency}
    \vspace{-5pt}
\end{wrapfigure}

\paragraph{Cross-benchmark perspective.}
The advantage of structured occlusion over simpler masking is much larger on RoboTwin. On RoboCasa, which already trains with randomized backgrounds, SDO-SC achieves 70.6\% versus 68.9\% for Bernoulli-SC, while Mixed Masking reaches 62.2\%.
% The advantage of structured occlusion over
% simpler masking is much larger on RoboTwin.
% On RoboCasa, SDO-SC achieves 70.6\%
% versus 68.9\% for Bernoulli-SC, while
% Mixed Masking reaches 62.2\%.
% Unlike RoboTwin's fixed-background training,
% RoboCasa already uses randomized backgrounds.
% Across both benchmarks, however, realistic
% natural-object textures provide no consistent
% additional benefit, and shared-color replacement
% outperforms independent coloring in aggregate.

\noindent\textbf{Finding 2: Spatial organization
matters beyond texture realism.}
On RoboTwin, SDO-SC substantially outperforms
simpler masking with shared-color replacement.
Adding realistic textures yields much smaller
differences and no consistent gain across
benchmarks, while shared colors outperform
independent coloring in aggregate.
These findings favor structured occlusion
geometry with simple, coherent replacement
appearances over more elaborate textures and colors.

\subsection{RQ3: Cross-View Allocation}
\label{sec:view_allocation}

We examine whether broader camera coverage
can compensate for weaker per-view corruption.
Using SDO, we compare single-view and
matched-budget masking at an equal summed
nominal budgets, alongside full-strength
all-view masking with a larger summed budget
(Sec.~\ref{sec:cross_view_design}).

\paragraph{Does broader coverage help at a fixed budget?}
On RoboTwin, single-view SDO achieves 53.4\% BG OOD success, compared
with 41.2\% when the same nominal budget is
distributed across all views
(Tab.~\ref{tab:robotwin}).
This 12.2 pp advantage has a 95\% task-bootstrap
CI of [7.3, 17.2] pp
(Appendix~\ref{app:statistics}).
On RoboCasa, the same ordering holds, but
with a much smaller difference
(68.3\% vs.\ 66.7\%;
Tab.~\ref{tab:robocasa}).
At a matched budget, spreading weaker masks across cameras is thus worse than concentrating them, clearly on RoboTwin and slightly on RoboCasa.

\paragraph{What if every view is strongly corrupted?}
Full-strength all-view SDO achieves the highest
aggregate success on both benchmarks.
On RoboTwin, it reaches 58.1\% BG OOD success,
exceeding single-view SDO by 4.7 pp
(95\% task-bootstrap CI: [1.4, 7.9])
and matched-budget SDO by 16.9 pp.
Relative to Standard, it improves success
on all 50 tasks
(Appendix Tab.~\ref{tab:app_robotwin_ood}).

On RoboCasa, full-strength all-view SDO
achieves 73.3\%, compared with 68.3\% for
single-view and 53.9\% for Standard.
The 5.0 pp gain over single-view has a
task-bootstrap CI of [-3.3, 13.9] pp.
Unlike matched-budget masking, full-strength
all-view SDO maintains the same occlusion
strength in each camera while increasing
the summed nominal budget.

\noindent\textbf{Finding 3: Broader view coverage
does not replace stronger per-view occlusion.}
At equal summed nominal budgets, concentrating
stronger corruption in one view outperforms
distributing weaker masks across views on
RoboTwin, with a smaller advantage on RoboCasa.
Full-strength all-view SDO achieves the highest
mean success on both benchmarks while using a larger summed budget.

% \begin{figure}[t]
%     \centering
%     \includegraphics[width=\columnwidth]{
%         img/fig3_d_bg_vs_ood_aggregate_inkteal_mm_clean.png
%     }
%     \caption{
%         \textbf{Background-induced action sensitivity
%         vs. OOD success.}
%         $D_{\mathrm{bg}}$ is measured on 10 randomly
%         selected RoboTwin tasks; success is aggregated
%         over all 50 tasks.
%         Horizontal error bars indicate the standard
%         error of $D_{\mathrm{bg}}$; the dashed line
%         shows a linear fit across 13 configurations.
%     }
%     \label{fig:dependency}
% \end{figure}

\begin{figure*}[t]
    \centering
    \includegraphics[width=\textwidth]{
        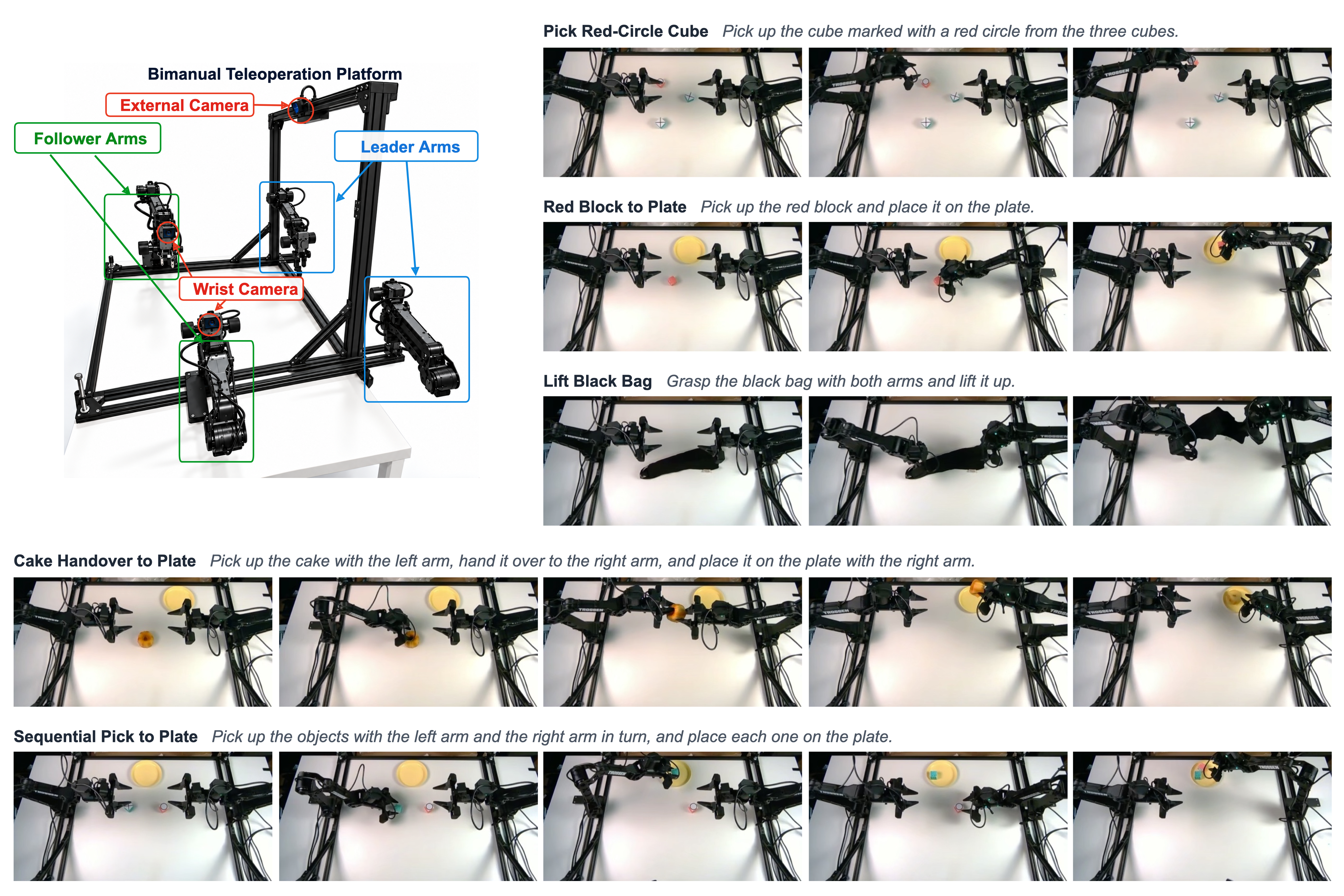
    }
    \caption{
        \textbf{Real-world bimanual manipulation tasks.}
        TROSSEN platform and representative execution
        sequences for Pick Red-Circle Cube, Red Block
        to Plate, Lift Black Bag, Cake Handover to Plate,
        and Sequential Pick to Plate.
    }
    \label{fig:real_robot_tasks}
\end{figure*}

% \subsection{Background Sensitivity and OOD Success}
% \label{sec:visual_dependency}

% Beyond task success, we examine whether
% robustness differences across corruption
% designs are reflected in how policies respond
% to task-irrelevant visual changes.
% We measure background-induced action
% sensitivity $D_{\mathrm{bg}}$
% (Eq.~\eqref{eq:bg_dependency}) by re-rendering
% identical robot and object states with different
% table and wall textures.
% Camera poses and nonvisual inputs remain fixed,
% and paired predictions use identical sampling
% noise. We evaluate 100 sampled states from
% 10 randomly selected RoboTwin tasks, using
% five background texture pairs and the first
% 24 predicted actions.

% \paragraph{Action sensitivity tracks OOD success.}
% Across 12 configurations with available
% $D_{\mathrm{bg}}$ measurements (excluding
% MAE-Matched), lower action sensitivity is
% strongly associated with higher background-OOD
% success (Fig.~\ref{fig:dependency};
% Pearson $r=-0.973$, Spearman $\rho=-0.972$).
% Standard exhibits high sensitivity
% ($D_{\mathrm{bg}}=0.247$) and only 7.6\%
% OOD success. In contrast, all-view SDO
% achieves 58.1\% success with substantially
% lower sensitivity ($D_{\mathrm{bg}}=0.056$).

% This strong association reveals a consistent
% behavioral pattern across corruption designs:
% policies with higher OOD success also produce
% more stable action predictions when only
% background appearance changes.

\subsection{Background Sensitivity and OOD Success}
\label{sec:visual_dependency}

Beyond task success, we examine how
corruption designs affect policies'
responses to task-irrelevant background
changes. 
We compute $D_{\mathrm{bg}}$
(Eq.~\ref{eq:bg_dependency}) on 100 states sampled from 10 randomly selected RoboTwin tasks, using five background textures and the first 24 predicted actions.
% We measure background-induced
% action sensitivity $D_{\mathrm{bg}}$
% (Eq.~\eqref{eq:bg_dependency}) by re-rendering
% identical robot and object states with
% different table and wall textures.
% Camera poses and nonvisual inputs remain
% fixed, and paired predictions use identical
% sampling noise. We evaluate 100 sampled states
% from 10 randomly selected RoboTwin tasks,
% using five background texture and
% the first 24 predicted actions.

\paragraph{Action sensitivity tracks OOD success.}
Across all 13 configurations, lower
$D_{\mathrm{bg}}$ strongly correlates with
higher BG OOD success over 50 tasks
(Fig.~\ref{fig:dependency};
Pearson $r=-0.972$, Spearman $\rho=-0.973$).
The association persists when success is
evaluated on the same 10 tasks
($r=-0.974$, $\rho=-0.956$)
and within the eight RQ2 configurations
($r=-0.933$, $\rho=-0.810$).
Consistent trends appear in key design
comparisons: SDO-SC reduces $D_{\mathrm{bg}}$
from 0.125 to 0.083 relative to Bernoulli-SC,
alongside a 15.8 pp OOD gain; single-view
SDO reduces it from 0.104 to 0.080 relative
to matched-budget SDO, alongside a 12.2 pp
gain. 
Interestingly, these OOD advantages do not
simply mirror clean success.
Bernoulli-SC exceeds SDO-SC on clean
evaluation (82.7\% vs.\ 80.6\%), while
matched-budget SDO slightly exceeds
single-view SDO (80.9\% vs.\ 79.2\%).
Both rankings reverse under BG OOD
(Tab.~\ref{tab:robotwin}).
Thus, background sensitivity aligns with
OOD robustness rather than clean-task
rankings in these comparisons.
These results associate stronger
OOD robustness with more stable action
predictions under background changes.
Complete measurements are reported in
Appendix~\ref{app:bg_sensitivity_detail}.

\subsection{Generality Across WAM Configurations}
\label{sec:wam_generality}

We further examine whether the benefits of
training-time partial visual evidence persist
across different WAM learning objectives.
Beyond our primary configuration, we consider
two representative learning formulations: joint
video--action modeling (\emph{Joint}) and
inverse-dynamics modeling (\emph{IDM}).
For each, we compare single-view SDO against
its corresponding Standard training baseline.

\begin{table}[t]
    \centering
    \footnotesize
    \setlength{\tabcolsep}{4pt}
    \renewcommand{\arraystretch}{1.06}

    % ==================== Table 4 ====================
    \begin{minipage}[t]{0.48\textwidth}
        \vspace{0pt}
        \centering
        \caption{
            \textbf{Generality across WAM configurations.}
            RoboTwin success (\%).
            $\Delta_{\mathrm{OOD}}$ is the BG OOD gain
            over the corresponding standard configuration (pp).
        }
        \label{tab:wam_generality}

        \begin{tabular}{@{}lccc@{}}
            \toprule
            Configuration & Clean & BG OOD
                & $\Delta_{\mathrm{OOD}}$ \\
            \midrule
            Joint       & 84.0 & 11.0 & -- \\
            Joint + SDO & 84.3 & \textbf{65.0}
                        & \textbf{+54.0} \\
            IDM         & 86.0 & 16.4 & -- \\
            IDM + SDO   & 86.8 & \textbf{63.8}
                        & \textbf{+47.4} \\
            \bottomrule
        \end{tabular}
    \end{minipage}
    \hfill
    % ==================== Table 5 ====================
    \begin{minipage}[t]{0.48\textwidth}
        \vspace{0pt}
        \centering
        \caption{
            \textbf{Real-robot success rates (\%).}
            Mean over five tasks (30 trials per task
            and condition). Occlusion denotes all-view
            SDO training. OOD Avg. covers four shifts;
            $\Delta$ denotes improvement in percentage
            points. Per-task results are in
            Appendix Tab.~\ref{tab:real_robot_full}.
        }
        \label{tab:real_robot}

        \begin{tabular}{@{}lccc@{}}
            \toprule
            Condition & Standard & Occlusion & $\Delta$ \\
            \midrule
            Clean       & 80.7 & 86.0 & +5.3 \\
            Blue BG     &  0.7 & 78.0 & +77.3 \\
            Brown BG    &  1.3 & 65.3 & +64.0 \\
            Low Light   & 10.0 & 72.7 & +62.7 \\
            Distractors & 49.3 & 56.0 & +6.7 \\
            \midrule
            \textbf{OOD Avg.}
                        & \textbf{15.3}
                        & \textbf{68.0}
                        & \textbf{+52.7} \\
            \bottomrule
        \end{tabular}
    \end{minipage}
\end{table}

% \begin{table}[t]
%     \centering
%     \footnotesize
%     \setlength{\tabcolsep}{4pt}
%     \caption{
%         \textbf{Generality across WAM configurations.}
%         RoboTwin success (\%); $\Delta_{\mathrm{OOD}}$
%         is the BG OOD gain over the corresponding
%         standard configuration (pp).
%     }
%     \label{tab:wam_generality}
%     \begin{tabular}{@{}lccc@{}}
%         \toprule
%         Configuration & Clean & BG OOD
%             & $\Delta_{\mathrm{OOD}}$ \\
%         \midrule
%         Joint       & 84.0 & 11.0 & -- \\
%         Joint + SDO & 84.3 & \textbf{65.0} & \textbf{+54.0} \\
%         \addlinespace[2pt]
%         IDM         & 86.0 & 16.4 & -- \\
%         IDM + SDO   & 86.8 & \textbf{63.8} & \textbf{+47.4} \\
%         \bottomrule
%     \end{tabular}
% \end{table}

\begin{figure}[t]
    \centering
    \begin{minipage}[t]{0.485\linewidth}
        \centering
        \includegraphics[width=\linewidth]{
            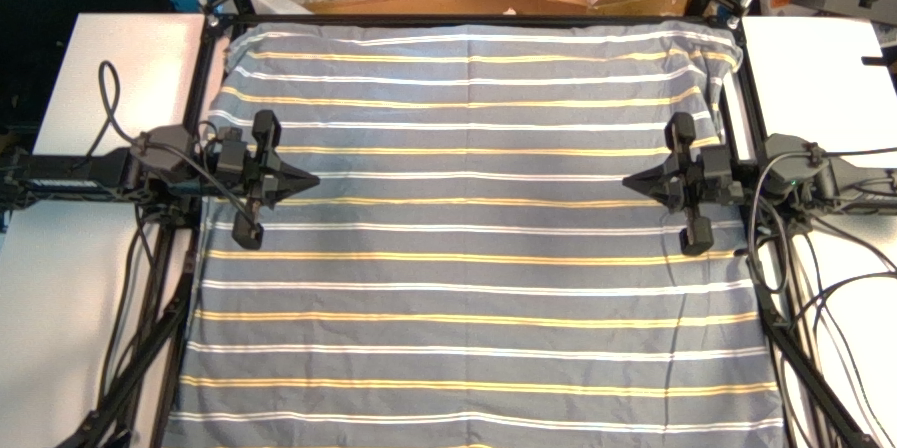
        }\\[-6pt]
        {\scriptsize (a) Blue BG}
    \end{minipage}\hfill
    \begin{minipage}[t]{0.485\linewidth}
        \centering
        \includegraphics[width=\linewidth]{
            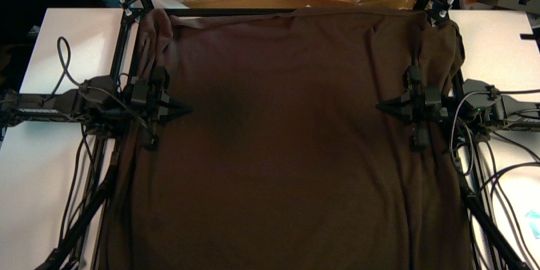
        }\\[-6pt]
        {\scriptsize (b) Brown BG}
    \end{minipage}

    \vspace{3pt}

    \begin{minipage}[t]{0.485\linewidth}
        \centering
        \includegraphics[width=\linewidth]{
            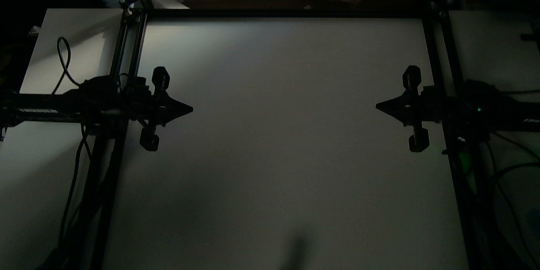
        }\\[-6pt]
        {\scriptsize (c) Low Light}
    \end{minipage}\hfill
    \begin{minipage}[t]{0.485\linewidth}
        \centering
        \includegraphics[width=\linewidth]{
            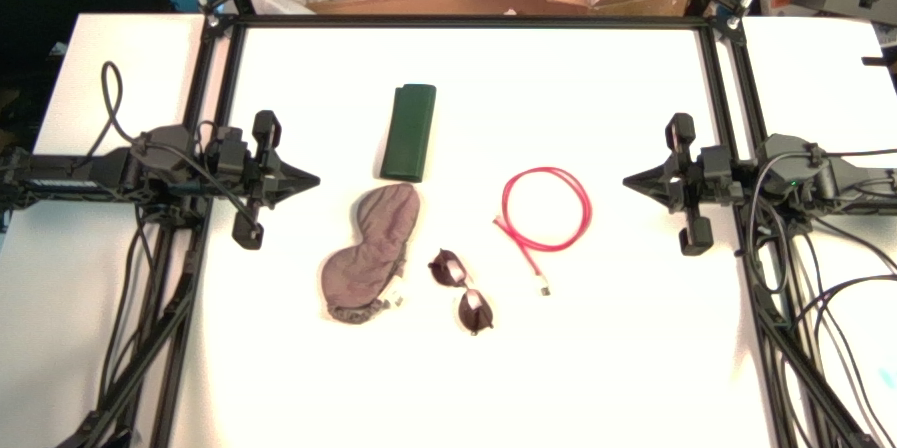
        }\\[-6pt]
        {\scriptsize (d) Distractors}
    \end{minipage}
    \caption{
        \textbf{Real-world visual shifts.}
        Two tabletop changes, reduced illumination,
        and task-irrelevant distractors.
    }
    \label{fig:real_robot_shifts}
\end{figure}

As shown in Tab.~\ref{tab:wam_generality},
SDO improves BG OOD success from 11.0\%
to 65.0\% in Joint (+54.0 pp) and from
16.4\% to 63.8\% in IDM (+47.4 pp).
Clean success is maintained or slightly
improved (84.0\% to 84.3\% for Joint;
86.0\% to 86.8\% for IDM).
Consequently, the clean-to-OOD performance
gap narrows from 73.0 to 19.3 pp for Joint
and from 69.6 to 23.0 pp for IDM.
The gains therefore primarily reflect
improved robustness to background shifts,
rather than higher clean-task success.

Despite differences in learning objectives
and baseline OOD performance, both
SDO-trained configurations reach similar
background-OOD success (65.0\% and 63.8\%).
The robustness benefit of training-time partial visual evidence thus holds for both training objectives.

\subsection{Real-Robot Validation}
\label{sec:real_robot}

We examine whether the robustness gains observed
in simulation extend to physical bimanual
manipulation under real visual shifts.
We compare standard Fast-WAM with its all-view
SDO-trained counterpart on a TROSSEN platform
equipped with two wrist cameras and one external
camera (Fig.~\ref{fig:real_robot_tasks}).
Both policies share training data, architecture,
and inference, differing only in training-time
visual corruption. 
% Test inputs remain unmasked.

\paragraph{Tasks and evaluation conditions.}
We evaluate five tasks spanning target selection,
placement, cooperative lifting, inter-arm handover,
and sequential manipulation
(Fig.~\ref{fig:real_robot_tasks}).
Each task is tested under the original training
appearance (Clean) and four shifts: two unseen
tabletop appearances (Blue BG and Brown BG),
reduced illumination (Low Light), and
task-irrelevant objects (Distractors;
Fig.~\ref{fig:real_robot_shifts}).
These conditions test robustness across changes
in surface appearance, illumination, and
surrounding visual clutter.
We conduct 30 trials per task, policy, and
condition, totaling 1,500 physical trials.
Task-level results and representative rollouts
are provided in Appendix~\ref{app:real_robot}.

% \begin{table}[t]
%     \centering
%     \footnotesize
%     \setlength{\tabcolsep}{4pt}
%     \renewcommand{\arraystretch}{1.06}
%     \caption{
%         \textbf{Real-robot success rates (\%).}
%         Mean over five tasks (30 trials per task
%         and condition). Occlusion denotes all-view
%         SDO training. OOD Avg. covers four shifts;
%         $\Delta$ denotes improvement in percentage
%         points. Per-task results are in
%         Appendix Tab.~\ref{tab:real_robot_full}.
%     }
%     \label{tab:real_robot}
%     \begin{tabular}{@{}lccc@{}}
%         \toprule
%         Condition & Standard & Occlusion & $\Delta$ \\
%         \midrule
%         Clean       & 80.7 & 86.0 & +5.3 \\
%         Blue BG     &  0.7 & 78.0 & +77.3 \\
%         Brown BG    &  1.3 & 65.3 & +64.0 \\
%         Low Light   & 10.0 & 72.7 & +62.7 \\
%         Distractors & 49.3 & 56.0 & +6.7 \\
%         \midrule
%         \textbf{OOD Avg.}
%                     & \textbf{15.3} & \textbf{68.0}
%                     & \textbf{+52.7} \\
%         \bottomrule
%     \end{tabular}
% \end{table}

\paragraph{Physical visual shifts expose a
large robustness gap.}
Standard Fast-WAM achieves 80.7\% success
under Clean but almost always fails under unseen
tabletop appearances (0.7\% and 1.3\%)
and drops to 10.0\% under Low Light
(Tab.~\ref{tab:real_robot}).
In contrast, all-view SDO achieves 78.0\%
under Blue BG, 65.3\% under Brown BG, and
72.7\% under Low Light, while increasing
Clean success to 86.0\%.
Across all four shifts, average success rises
from 15.3\% to 68.0\% (+52.7 pp).
The largest gains occur under the two unseen
backgrounds (+77.3 and +64.0 pp), with a
similar gain under
reduced illumination (+62.7 pp).
Thus, the robustness gains extend beyond
the background-texture changes used in the
primary simulation evaluation.
The gap between clean and average shifted
success decreases from 65.4 pp for Standard
to 18.0 pp for all-view SDO.
While clean success improves by only 5.3 pp,
the average gain across visual shifts reaches
52.7 pp.
This contrast highlights the contribution
of corruption training to preserving task
performance under visual distribution shifts.

\paragraph{Cross-task consistency and shift variation.}
The gains span distinct bimanual manipulation
skills rather than a few favorable tasks.
Under Blue BG, Lift Black Bag improves from
0/30 to 28/30 successes, while Cake Handover
improves from 1/30 to 26/30 under Brown BG
(Appendix Tab.~\ref{tab:real_robot_full}).
Across Blue BG, Brown BG, and Low Light,
SDO training improves all 15 task-condition
pairs. Clean success also improves on all
five tasks, indicating that the OOD gains
do not come at the expense of clean success
in this evaluation.

The magnitude of improvement differs across
visual shifts. Under Distractors, average
success rises from 49.3\% to 56.0\%, a smaller
gain than under backgrounds or low light.
Three tasks improve, one ties, and one declines;
Sequential Pick drops from 20/30 to 17/30
successes in this condition.
Some background-shift cases also remain
challenging: Pick Red-Circle Cube achieves
9/30 successes under Brown BG.
These results reveal strong, broad gains under
appearance shifts alongside more variable
responses to additional distractor objects.

\noindent\textbf{Real-world observation.}
Training-time partial visual evidence transfers
substantial robustness gains to physical
bimanual manipulation, raising average success
across four visual shifts from 15.3\% to 68.0\%.
The improvements span all five tasks under
unseen backgrounds and reduced illumination,
while gains under distractors are smaller
and more task-dependent. Together, these
results demonstrate robust physical transfer
without modifying test-time observations
or inference.

\section{Conclusion}
\label{sec:conclusion}

We studied how the representation,
within-view design, and cross-view allocation
of missing visual evidence shape WAM robustness.
Three findings emerge.
First, nominally aligned masking does not imply
equivalent robustness, with pixel-space corruption
achieving higher aggregate success than latent-token
masking on both benchmarks.
Second, spatial organization matters beyond
texture realism: SDO-SC outperforms simpler
masking on RoboTwin and performs comparably
to textured SDO on both benchmarks, while
shared colors outperform independent coloring.
Third, broader view coverage does not replace
stronger per-view occlusion: at equal summed
nominal budgets, concentrating corruption
in one view performs better on RoboTwin,
while full-strength all-view SDO achieves
the highest aggregate success on both benchmarks.

The advantages of structured occlusion and
concentrated masking are larger on
fixed-background RoboTwin than on
background-randomized RoboCasa.
Lower background-induced action sensitivity
tracks higher OOD success, while all-view SDO
raises mean real-robot success across four
visual shifts from 15.3\% to 68.0\%.
These findings provide practical guidance
for training-time visual corruption in WAMs.

\paragraph{Limitations.}
Our experiments focus on one WAM model
family, with a single training run per
condition. Task-bootstrap intervals capture
variation across tasks rather than training
seeds. While our results highlight the
importance of spatial occlusion design,
they do not isolate the contributions of
individual geometric properties, such as
region connectivity, scale distribution,
and spatial arrangement. Understanding
which properties drive robustness gains
remains an important direction for
future research.

% Bibliography
% \newpage
\bibliographystyle{plainnat}
\bibliography{references}

\clearpage
\setcounter{page}{1}
% \maketitlesupplementary

\appendix

\section{Implementation and Training Details}
\label{app:implementation}

This section provides additional details on the
world-action model implementation, optimization,
checkpoint selection, evaluation seeds, and
computational resources used in our study.

\subsection{WAM Architecture and Training Objective}
\label{app:wam_architecture}

\paragraph{Backbone architecture.}
Our main experiments build on
Fast-WAM~\cite{yuan2026fast}, using a
mixture-of-transformers (MoT) architecture
with separate video and action experts.

The video expert is initialized from the
Wan2.2-TI2V-5B diffusion transformer (DiT).
It contains 30 transformer layers, a hidden
dimension of 3072, 24 attention heads,
and a feed-forward dimension of 14336.
Its spatiotemporal patch size is
$1\times2\times2$.

The action expert, denoted ActionDiT,
contains 30 transformer layers, a hidden
dimension of 1024, 24 attention heads,
and a feed-forward dimension of 4096.
Its transformer weights are initialized
from Wan2.2-derived weights using
linear interpolation and scaling.
The action input/output projections and
proprioceptive projection are newly
initialized.
The two experts jointly process video and
action tokens through the MoT architecture.

\paragraph{Frozen and trainable components.}
We use a frozen Wan2.2 video variational
autoencoder (VAE).
Language conditioning uses precomputed
umT5 embeddings with a dimension of 4096
and a maximum sequence length of 128 tokens.
The text encoder is not loaded or updated
during policy training.

Both the video and action transformer experts,
together with the proprioceptive projection,
are fully optimized.

\paragraph{Conditioning and attention.}
The encoded current observation is inserted
as the first latent video frame, following
the TI2V-style conditioning scheme.
Language embeddings provide cross-attention
context, while the proprioceptive state at
the first step is projected into a single
token and appended to the conditioning context.

Video tokens follow a first-frame-causal
attention pattern, while action tokens
attend to one another.
Among video tokens, action tokens attend
only to the current conditioning frame;
video tokens do not attend to action tokens.

Consequently, actions can be generated
directly at inference time without first
synthesizing future video frames.

\paragraph{Joint flow-matching objective.}
Both the video and action experts are trained
using flow matching.
For a clean video-latent or action target
$x_0$ and Gaussian noise $\epsilon$,
the velocity target is
\begin{equation}
    v^\star = \epsilon - x_0.
    \label{eq:app_flow_target}
\end{equation}

Video and action flow-matching time steps
are sampled independently.
For $k\in\{v,a\}$, we sample
$u_k\sim\mathcal{U}(0,1)$ and define
\begin{equation}
    t_k = 1000\,
    \frac{5u_k}{1+4u_k}.
    \label{eq:app_timestep}
\end{equation}
This corresponds to a time-step shift of 5
over 1000 training time steps.

We follow the time-dependent loss weighting
scheme of Fast-WAM~\cite{yuan2026fast}.
The video and action loss coefficients
are set to
$\lambda_v=\lambda_a=1$.
All predicted future latent frames contribute
equally to the video loss.

\paragraph{Corrupted conditioning and clean targets.}
For pixel-space corruption, the augmentation
is applied to the current visual observations
before VAE encoding.
Only the conditioning pathway is modified:
future-video targets are encoded from the
original uncorrupted observations.

The video loss excludes the first latent
frame and is computed only on predicted
future latent frames.
Thus, the model is not trained to reconstruct
the artificially corrupted conditioning frame.
No additional masked-image reconstruction
objective is introduced.

For the latent-masking condition in RQ1,
visual evidence is instead suppressed through
attention masking of current-frame latent
tokens, without modifying the input RGB
images or adding trainable parameters.
The underlying action and future-video
supervision remains unchanged.

Additional details of the corruption
implementations are provided in
Appendix~\ref{app:visual_corruption}.

\subsection{Training Hyperparameters}
\label{app:training_details}

\paragraph{Shared optimization settings.}
Unless otherwise specified, conditions within
each benchmark use the same model backbone,
training demonstrations, loss coefficients,
optimization schedule, and inference procedure.
The experimental interventions affect only
the training-time visual conditioning.

Table~\ref{tab:app_shared_hparams}
summarizes the shared optimization settings.

\begin{table}[t]
    \centering
    \caption{
        \textbf{Shared training hyperparameters.}
        Settings common to the RoboTwin and
        RoboCasa experiments.
    }
    \label{tab:app_shared_hparams}
    \footnotesize
    \setlength{\tabcolsep}{4pt}
    \renewcommand{\arraystretch}{1.08}
    \begin{tabular}{
        @{}p{0.40\columnwidth}
        p{0.51\columnwidth}@{}
    }
        \toprule
        Hyperparameter & Value \\
        \midrule
        Optimizer & AdamW \\
        AdamW $\beta_1,\beta_2$ & $0.9,\ 0.95$ \\
        Weight decay & $0.01$ \\
        Gradient clipping & $1.0$ \\
        Peak learning rate & $1\times10^{-4}$ \\
        Final learning rate & $1\times10^{-6}$ \\
        LR schedule & Linear warmup + cosine decay \\
        Warmup fraction & $5\%$ \\
        Global batch size & $128$ \\
        Numerical precision & bfloat16 \\
        Distributed optimization & DeepSpeed ZeRO-1 \\
        Video / action loss weights & $1.0 / 1.0$ \\
        Flow time-step shift & $5$ \\
        Training time steps & $1000$ \\
        Action prediction horizon & $32$ steps \\
        Video training frames & $9$ (1 current + 8 future) \\
        Future-frame stride & $4$ action steps \\
        Inference solver & Euler \\
        Inference solver steps & $10$ \\
        Classifier-free guidance & Disabled (scale $1.0$) \\
        \bottomrule
    \end{tabular}
\end{table}

The learning-rate scheduler is updated once
per optimizer step.
For RoboTwin, the learning rate reaches
$10^{-4}$ after 1500 warmup steps and
decreases to $10^{-6}$ at the final step.
The same peak and final learning rates are
used for RoboCasa, with a longer schedule.

\paragraph{Benchmark-specific settings.}
Table~\ref{tab:app_benchmark_hparams}
reports the main settings that differ
between RoboTwin and RoboCasa.

\begin{table}[t]
    \centering
    \caption{
        \textbf{Benchmark-specific configurations.}
        Training data, observation layout,
        optimization duration, and action execution.
    }
    \label{tab:app_benchmark_hparams}
    \footnotesize
    \setlength{\tabcolsep}{2.5pt}
    \renewcommand{\arraystretch}{1.10}
    \begin{tabular}{
        @{}p{0.22\columnwidth}
        p{0.35\columnwidth}
        p{0.35\columnwidth}@{}
    }
        \toprule
        Setting & RoboTwin & RoboCasa \\
        \midrule
        Tasks
        & 50
        & 18 \\

        Demonstrations
        & 2,500
        & 9,126 \\

        Data sampling
        & Task-balanced
        & Task-balanced \\

        Camera layout
        & 1 head + 2 wrists
        & 3 image views \\

        Image layout
        & Head: $256\times320$;
          wrists: $128\times160$ each
        & Three $224\times224$ views;
          horizontal concatenation \\

        Action / proprio dimensions
        & $14/14$
        & $12/16$ \\

        Normalization
        & Z-score
        & Min-max \\

        Training steps
        & 30,000
        & 52,299 \\

        Warmup steps
        & 1,500
        & 2,614 \\

        GPUs
        & 8
        & 8 \\

        Gradient accumulation
        & 1
        & 2 \\

        Actions executed
        per replan
        & 24
        & 10 \\

        \bottomrule
    \end{tabular}
\end{table}

\paragraph{Visual observation layout.}
For RoboTwin, the head-camera image is
resized to $256\times320$ and placed above
the two wrist-camera images, each resized
to $128\times160$ and arranged side by side.

For RoboCasa, three $224\times224$ images
are concatenated horizontally, resulting
in a $224\times672$ observation canvas.

The action and proprioceptive dimensions
for both benchmarks are listed in
Tab.~\ref{tab:app_benchmark_hparams}.
RoboTwin uses Z-score normalization,
whereas RoboCasa uses min-max normalization.

\paragraph{Temporal sampling and inference.}
Training uses a 33-step temporal window
containing 32 action steps.
Video supervision contains nine frames:
the current frame and eight future frames
sampled at intervals of four action steps.

At inference, the policy predicts
32-action chunks using a 10-step Euler
solver without classifier-free guidance.
For RoboTwin, the policy executes
24 predicted actions before replanning.
For RoboCasa, it executes 10 actions
before replanning, at a control
frequency of 20 Hz.

\paragraph{Configuration-specific implementation.}
The IDM-style configurations use gradient
checkpointing to reduce memory consumption.
This changes the memory--computation tradeoff
but not the mathematical training objective.

\subsection{Checkpoint Selection and Random Seeds}
\label{app:checkpoint_seeds}

\paragraph{Checkpoint selection.}
All results use the final training checkpoint.
For RoboTwin, this corresponds to
optimizer step 30,000; for RoboCasa,
the saved final checkpoint is at step 52,299.

No checkpoint is selected based on
evaluation success rates.
We do not perform early stopping or
intermediate validation-based model selection.

\paragraph{Training randomness.}
Each reported experimental condition
corresponds to a single independent training run.

The training configuration uses a base
random seed of 42.
Data-order and flow-noise randomness
use process-dependent seeds derived
from the base seed and distributed rank.
% For runs supporting explicit model-initialization
% seeding, newly initialized parameters are
% also seeded with 42.

\paragraph{RoboTwin evaluation.}
We use three matched evaluation seed groups,
indexed by $g\in\{0,1,2\}$.
The scene-seed search for group $g$ starts at
\begin{equation}
    s_g=100000(1+g).
    \label{eq:app_robotwin_seed}
\end{equation}

Candidate scene seeds are checked against
the expert-success cache, and 10 valid
initializations are selected for each task
in each evaluation group.
The action-sampling noise seed is set to
the group index $g$.

All compared models within a group use
the same selected scene initializations.
With 50 tasks and 10 episodes per task,
each group contains 500 evaluation rollouts.
Thus, the standard RoboTwin evaluation
comprises 1,500 rollouts per model
and visual condition across three groups.

A small number of scene seeds that triggered
renderer failures were replaced with subsequent
valid seeds.
We report aggregate success rates over
30 evaluation episodes per task.

\paragraph{RoboCasa evaluation.}
RoboCasa uses 18 tasks with 10 evaluation
episodes per task, for 180 episodes per model.

For task index $i\in\{0,\ldots,17\}$ and
repetition index $k\in\{0,\ldots,9\}$,
the episode seed is
\begin{equation}
    s_{i,k}=2026+18k+i.
    \label{eq:app_robocasa_seed}
\end{equation}

Policy sampling noise is initialized
using the episode seed together with
the current replanning index.
Episode horizons follow the official
task-specific RoboCasa settings.

Each RoboCasa model is evaluated using
one evaluation seed set. 
We report aggregate success rates over
10 episodes per task.

\paragraph{Interpretation of uncertainty.}
The evaluation protocols above quantify
performance over matched task initializations
and evaluation randomness.
Where paired task-bootstrap confidence
intervals are reported, their sampling
unit is the task, and their interpretation
is described in Appendix~\ref{app:statistics}.

\subsection{Computational Resources and Cost}
\label{app:compute}

\paragraph{Compute infrastructure.}
Training is conducted on GPU clusters
using eight NVIDIA A100 80GB GPUs per run.
Depending on cluster availability, each run
uses either a single eight-GPU node or
two four-GPU nodes connected through
InfiniBand.

The numerical training precision is
bfloat16, with DeepSpeed ZeRO-1 used
for distributed optimization.

\paragraph{Training cost.}
Table~\ref{tab:app_compute_cost}
summarizes the approximate wall-clock
training cost observed for the completed
experiment groups.
Wall-clock time includes initialization,
preflight checks, training, and checkpoint
saving.

\begin{table}[t]
    \centering
    \caption{
        \textbf{Approximate training cost per run.}
        All configurations use eight NVIDIA A100 GPUs.
        GPU-hours are computed from job wall-clock
        duration and allocated GPU count.
    }
    \label{tab:app_compute_cost}
    \footnotesize
    \setlength{\tabcolsep}{3pt}
    \renewcommand{\arraystretch}{1.08}
    \begin{tabular}{
        @{}p{0.48\columnwidth}
        p{0.18\columnwidth}
        p{0.22\columnwidth}@{}
    }
        \toprule
        Training group
        & Hours
        & GPU-hours \\
        \midrule

        RoboTwin main configurations
        & 65.6--66.2
        & 525--530 \\

        RoboTwin Joint / IDM
        & 65.4--66.6
        & 524--533 \\

        RoboTwin shared-color controls
        & 50.5--50.9
        & $\sim405$ \\

        RoboTwin Mixed Masking
        & 50.8
        & $\sim407$ \\

        \midrule

        RoboCasa main configurations
        & 42.1--47.9
        & 337--383 \\

        RoboCasa shared-color controls
        & 79.0--79.3
        & $\sim633$ \\

        RoboCasa Mixed Masking
        & 48.6
        & $\sim388$ \\

        \bottomrule
    \end{tabular}
\end{table}

% \paragraph{Total training compute.}
Across 30 training runs spanning RoboTwin,
RoboCasa, and additional WAM configurations,
the cumulative training cost is approximately
13,800--14,300 A100 GPU-hours.
GPU-hours are estimated from allocated GPU
counts and job wall-clock durations, including
initialization and checkpoint saving.
Evaluation costs are excluded.

The observed wall-clock differences reflect
the hardware allocation and runtime behavior
of different cluster configurations,
rather than changes to the underlying
optimization budgets.

\paragraph{Evaluation cost.}
RoboTwin evaluations use four GPUs.
A standard evaluation group contains
500 rollouts; clean-background evaluation
typically requires approximately
5.5--5.9 hours per group.

RoboCasa evaluation uses four GPUs
and requires approximately one hour
for 180 episodes.

% \paragraph{Software versions.}
% The training environment uses
% Python 3.10.20, PyTorch 2.7.1,
% CUDA 12.8, cuDNN 9.7,
% Accelerate 1.12.0, and Hydra 1.3.2.

% RoboTwin evaluation uses
% SAPIEN 3.0.0b1 and mplib 0.2.1,
% with the RoboTwin codebase at
% commit \texttt{45d8e14}.

% RoboCasa evaluation uses
% RoboCasa 1.0.1, robosuite 1.5.2,
% and MuJoCo 3.3.1.

\begin{figure*}[t]
    \centering
    \includegraphics[width=\textwidth]{
        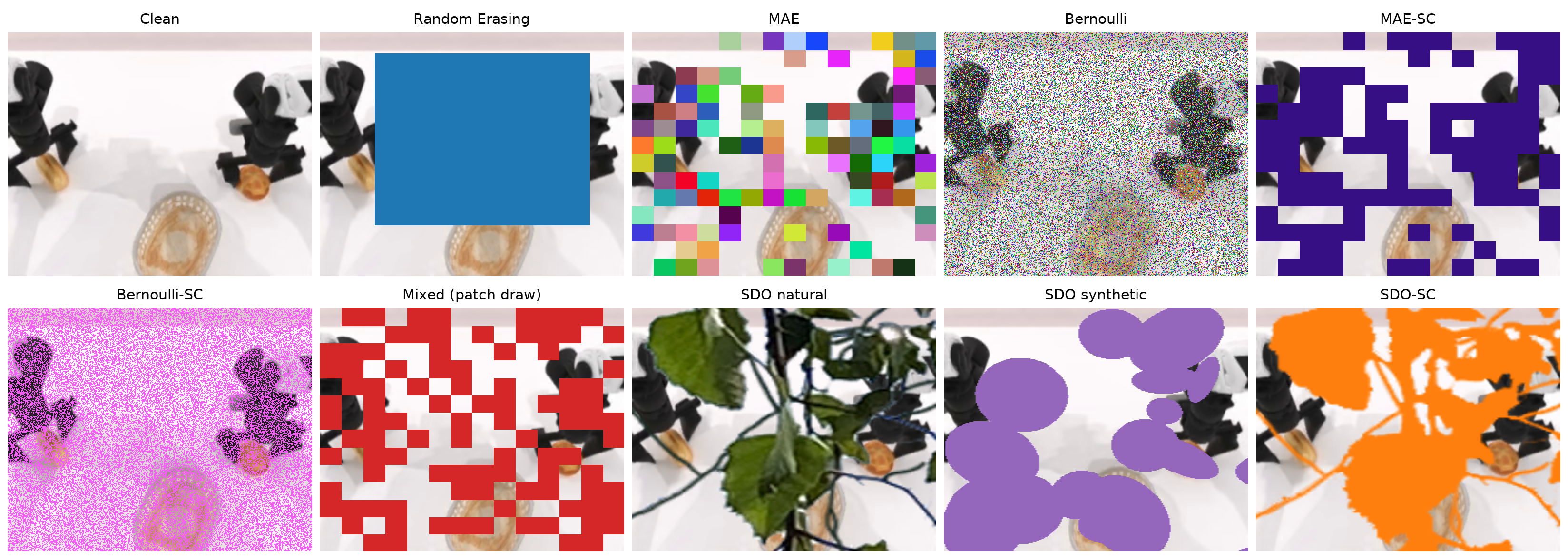
    }
    \caption{
        \textbf{Within-view corruption examples.}
        Different corruption schemes applied to
        the same clean RoboTwin observation.
        A nominal ratio $\rho=0.5$ is used
        for illustration; training samples
        $\rho$ uniformly from $[0,1]$.
        Mixed Masking shows one sampled patch-based
        realization.
    }
    \label{fig:app_within_view}
\end{figure*}

\begin{figure*}[t]
    \centering
    \includegraphics[width=\textwidth]{
        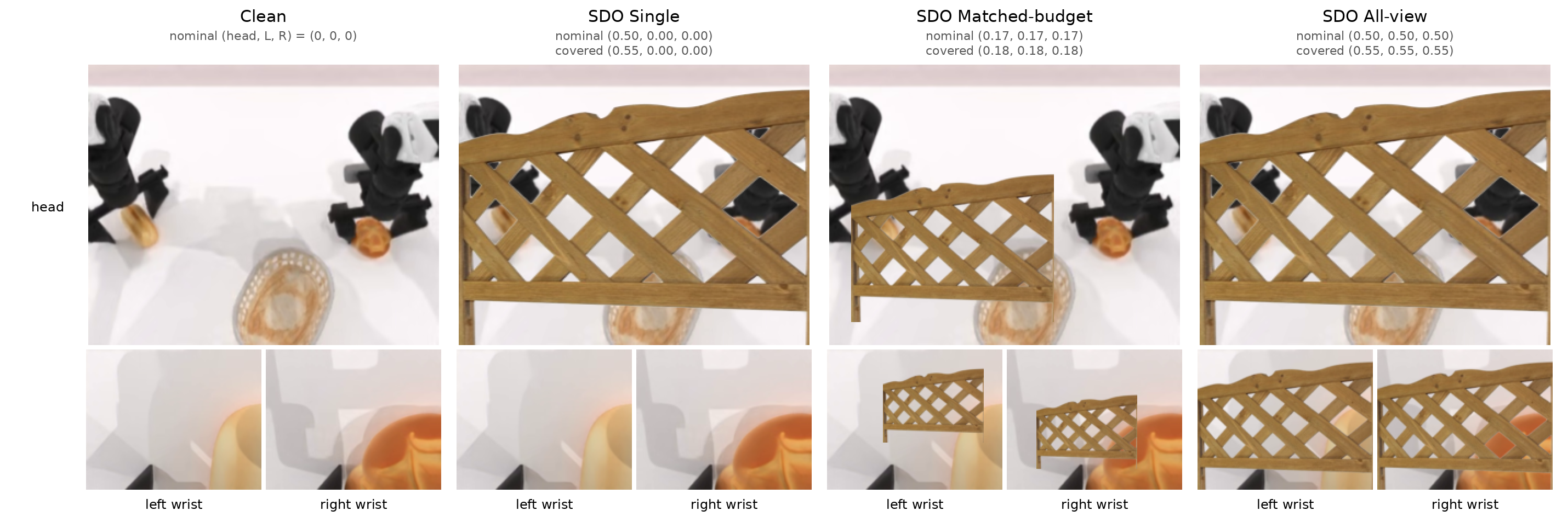
    }
\caption{
    \textbf{Cross-view corruption examples.}
    SDO applied to three camera views with
    a sampled ratio of $\rho=0.5$.
    Single-view corrupts one camera;
    matched-budget distributes the nominal
    budget across views using different
    placements; all-view applies full-strength
    corruption with shared placement.
}
    \label{fig:app_cross_view}
\end{figure*}
% ============================================================
% Appendix B: Visual Corruption
% ============================================================

\section{Training-Time Visual Corruption}
\label{app:visual_corruption}

This section provides additional details and visual
examples of the training-time corruption schemes
used in our study.

\subsection{Common Corruption Sampling}

For pixel-space conditions other than
MAE-Matched, each training sample remains
clean with probability $p_{\mathrm{clean}}=0.5$.
Otherwise, we sample a masking ratio
$\rho\sim\mathcal{U}(0,1)$ and apply the
selected corruption. These conditions share
the same sampling protocol.

For RQ1, MAE-Matched and latent masking use
the same clean probability and masking-ratio
distribution $q\sim\mathcal{U}(0,1/3)$.
Both sample mask locations uniformly across
the three-view observation canvas.
MAE-Matched replaces selected $32\times32$
RGB patches with a single randomly sampled
color shared across all masked patches.
The patch size corresponds to the nominal
image footprint of one visual token, given
the VAE's $16\times$ spatial downsampling
and the DiT's $2\times2$ spatial patching.
Latent masking instead suppresses selected
visual tokens through attention after encoding.

All corruptions affect only training-time
visual conditioning; action and future-video
targets remain clean, and evaluation
observations are always unmasked.

\subsection{Within-View Corruption Schemes}

Figure~\ref{fig:app_within_view} illustrates
the within-view corruption schemes applied
to the same RoboTwin observation.
Random Erasing replaces a rectangular region
with a solid color. MAE-style masking selects
random patches on a $14\times14$ grid, while
Bernoulli masking independently samples
masked pixels. Their shared-color variants,
MAE-SC and Bernoulli-SC, preserve the same
mask geometry but use a single randomly
sampled RGB color per training sample,
rather than independent colors per patch
or pixel. Mixed Masking uniformly selects
among Random Erasing, MAE-SC, and
Bernoulli-SC.

Structured Diverse Occlusion (SDO) samples
natural-object cutouts and synthetic occlusion
patterns with equal probability.
Natural-object cutouts retain their RGB
textures, while synthetic patterns use
randomly colored geometric structures.
SDO-SC follows the same shared-color
principle as MAE-SC and Bernoulli-SC,
replacing natural-object textures with
a single random RGB color per sample.
It preserves SDO's occlusion geometry,
sampling procedure, and synthetic branch.

\subsection{Cross-View Corruption Allocation}

Figure~\ref{fig:app_cross_view} illustrates
three SDO allocation strategies across
one head camera and two wrist cameras.
Single-view corruption applies ratio $\rho$
to one uniformly selected camera, leaving
the other two unchanged.
Matched-budget corruption distributes the
same summed nominal masking budget across
all three cameras, applying $\rho/3$ to
each view.

In contrast, full-strength all-view corruption
applies ratio $\rho$ to every camera. Its summed nominal masking
budget is therefore $3\rho$, compared with
$\rho$ for single-view and matched-budget
corruption. The first two strategies compare
cross-view allocation at equal summed nominal
budgets, while the third evaluates stronger
per-view corruption across all cameras.

% ============================================================
% Appendix C: Simulation Benchmarks Tasks Analysis
% ============================================================

% ============================================================
% Appendix C: Task-Level Simulation Results and Analysis
% ============================================================
% \clearpage

\section{Simulation Results and Task-Level Analysis}
\label{app:simulation_results}
\label{app:benchmarks}

We report complete task-level results for RoboTwin
and RoboCasa and examine how aggregate differences
vary across tasks. Training and evaluation settings
are described in Sec.~3.3 and Appendix~A.

\subsection{RoboTwin Task-Level Results}
\label{app:robotwin_results}

Tables~\ref{tab:app_robotwin_ood} and
\ref{tab:app_robotwin_clean} report per-task
success counts under background-OOD and clean
conditions, respectively, with 30 trials per task.
Column abbreviations are: Std, standard training;
Lat, latent-token masking; MM, MAE-Matched;
MAE, MAE-style masking; RE, Random Erasing;
Ber, Bernoulli; M-SC and B-SC, their shared-color
variants; Mix, Mixed Masking; S-SC, SDO-SC;
and S-S, S-M, and S-A, single-view,
matched-budget, and all-view SDO.

For RQ1, MAE-Matched outperforms latent masking
on 27 tasks, ties on 14, and performs worse
on nine, yielding a 3.6 pp aggregate advantage.

For RQ2, SDO-SC outperforms Bernoulli-SC
on 42 tasks, ties on three, and performs
worse on five, indicating that the advantage
of structured occlusion extends across
most RoboTwin tasks.
Shared-color MAE and Bernoulli masking also
outperform their independent-color counterparts
on 34 and 30 tasks, respectively.
By comparison, textured SDO outperforms
SDO-SC on 27 tasks, ties on seven, and
performs worse on 16, with a much smaller
aggregate difference of 2.8 pp.

For RQ3, single-view SDO outperforms
matched-budget masking on 34 tasks,
ties on four, and performs worse on 12.
Full-strength all-view SDO improves
upon Standard on all 50 tasks, while
outperforming single-view SDO on 27 tasks,
tying on seven, and performing worse on 16.
Thus, its large overall robustness gain
is broadly distributed across tasks,
although its advantage over single-view
corruption varies by task.

\begin{table}[t]
    \centering
    \caption{
        \textbf{Background sensitivity and
        BG OOD success on RoboTwin.}
        $D_{\mathrm{bg}}$ is measured on
        100 states from 10 tasks.
        $S_{50}$ denotes success over all
        50 tasks, while $S_{10}$ uses the
        same 10 tasks as $D_{\mathrm{bg}}$.
        Success rates are in percent.
    }
    \label{tab:app_bg_sensitivity}
    \footnotesize
    \setlength{\tabcolsep}{3pt}
    \renewcommand{\arraystretch}{1.08}
    \begin{tabular}{@{}lccc@{}}
        \toprule
        Method
        & $D_{\mathrm{bg}}\pm\mathrm{SE}$
        & $S_{50}$
        & $S_{10}$ \\
        \midrule

        Standard
        & $0.247\pm0.012$ & 7.6 & 8.3 \\

        \midrule
        Latent
        & $0.186\pm0.013$ & 17.0 & 25.0 \\
        MAE-Matched
        & $0.198\pm0.011$ & 20.6 & 24.0 \\

        \midrule
        Random Erasing
        & $0.138\pm0.017$ & 27.4 & 34.3 \\
        MAE-style
        & $0.168\pm0.014$ & 22.2 & 30.3 \\
        Bernoulli
        & $0.156\pm0.016$ & 29.5 & 42.3 \\
        MAE-SC
        & $0.132\pm0.017$ & 30.1 & 41.0 \\
        Bernoulli-SC
        & $0.125\pm0.017$ & 34.8 & 46.0 \\
        Mixed Masking
        & $0.133\pm0.014$ & 36.9 & 50.3 \\
        SDO-SC
        & $0.083\pm0.015$ & 50.6 & 68.3 \\
        SDO (single)
        & $0.080\pm0.014$ & 53.4 & 67.7 \\

        \midrule
        SDO (matched)
        & $0.104\pm0.015$ & 41.2 & 59.7 \\
        SDO (all)
        & $0.056\pm0.014$ & 58.1 & 73.0 \\

        \bottomrule
    \end{tabular}
\end{table}

\begin{table*}[p]
    \centering
    \footnotesize
    \setlength{\tabcolsep}{2.6pt}
    \renewcommand{\arraystretch}{0.97}
    \caption{
        \textbf{Complete RoboTwin background-OOD results.}
        Successful trials out of 30 for each task and
        training condition.
        The final row gives aggregate success rates (\%).
    }
    \label{tab:app_robotwin_ood}
    \begin{tabular}{@{}l*{13}{c}@{}}
        \toprule
        & & \multicolumn{2}{c}{RQ1}
          & \multicolumn{8}{c}{RQ2}
          & \multicolumn{2}{c}{RQ3} \\
        \cmidrule(lr){3-4}
        \cmidrule(lr){5-12}
        \cmidrule(l){13-14}
        Task & Std & Lat & MM & MAE & RE & Ber
             & M-SC & B-SC & Mix & S-SC & S-S & S-M & S-A \\
        \midrule
        \texttt{adjust\_bottle} & 3 & 8 & 13 & 17 & 14 & 15 & 15 & 24 & 17 & 22 & 29 & 24 & 28 \\
        \texttt{beat\_block\_hammer} & 0 & 5 & 2 & 5 & 6 & 10 & 5 & 8 & 11 & 19 & 12 & 15 & 20 \\
        \texttt{blocks\_ranking\_rgb} & 0 & 2 & 1 & 1 & 4 & 7 & 9 & 5 & 14 & 19 & 14 & 24 & 20 \\
        \texttt{blocks\_ranking\_size} & 0 & 0 & 1 & 1 & 2 & 2 & 2 & 2 & 3 & 2 & 2 & 2 & 2 \\
        \texttt{click\_alarmclock} & 18 & 10 & 10 & 14 & 15 & 10 & 18 & 18 & 18 & 22 & 27 & 14 & 29 \\
        \texttt{click\_bell} & 2 & 1 & 4 & 4 & 3 & 7 & 7 & 8 & 10 & 21 & 25 & 6 & 28 \\
        \texttt{dump\_bin\_bigbin} & 2 & 5 & 5 & 4 & 12 & 6 & 7 & 9 & 17 & 18 & 23 & 16 & 30 \\
        \texttt{grab\_roller} & 8 & 18 & 20 & 20 & 22 & 22 & 22 & 24 & 23 & 29 & 29 & 25 & 29 \\
        \texttt{handover\_block} & 0 & 1 & 1 & 3 & 1 & 1 & 1 & 2 & 3 & 10 & 13 & 8 & 16 \\
        \texttt{handover\_mic} & 0 & 1 & 2 & 1 & 9 & 4 & 8 & 7 & 6 & 20 & 22 & 18 & 25 \\
        \texttt{hanging\_mug} & 0 & 0 & 0 & 0 & 0 & 2 & 2 & 0 & 3 & 4 & 3 & 0 & 4 \\
        \texttt{lift\_pot} & 1 & 3 & 5 & 8 & 10 & 11 & 9 & 13 & 11 & 23 & 25 & 14 & 23 \\
        \texttt{move\_can\_pot} & 0 & 1 & 3 & 6 & 2 & 7 & 4 & 5 & 7 & 20 & 17 & 7 & 21 \\
        \texttt{move\_pillbottle\_pad} & 1 & 2 & 5 & 4 & 6 & 5 & 9 & 15 & 10 & 20 & 26 & 16 & 20 \\
        \texttt{move\_playingcard\_away} & 1 & 6 & 10 & 7 & 8 & 12 & 8 & 12 & 15 & 18 & 13 & 14 & 18 \\
        \texttt{move\_stapler\_pad} & 1 & 1 & 1 & 0 & 1 & 2 & 2 & 3 & 2 & 3 & 4 & 3 & 5 \\
        \texttt{open\_laptop} & 2 & 9 & 7 & 15 & 15 & 13 & 13 & 19 & 14 & 21 & 26 & 17 & 24 \\
        \texttt{open\_microwave} & 1 & 0 & 1 & 1 & 1 & 3 & 1 & 2 & 9 & 4 & 12 & 0 & 12 \\
        \texttt{pick\_diverse\_bottles} & 3 & 6 & 6 & 7 & 9 & 5 & 10 & 8 & 8 & 7 & 10 & 11 & 8 \\
        \texttt{pick\_dual\_bottles} & 2 & 7 & 6 & 15 & 11 & 8 & 13 & 5 & 7 & 10 & 16 & 17 & 19 \\
        \texttt{place\_a2b\_left} & 1 & 3 & 3 & 2 & 3 & 7 & 5 & 8 & 6 & 9 & 10 & 8 & 6 \\
        \texttt{place\_a2b\_right} & 1 & 3 & 3 & 2 & 3 & 4 & 2 & 7 & 4 & 8 & 9 & 5 & 7 \\
        \texttt{place\_bread\_basket} & 0 & 9 & 10 & 5 & 6 & 12 & 13 & 11 & 13 & 15 & 11 & 15 & 13 \\
        \texttt{place\_bread\_skillet} & 1 & 2 & 2 & 3 & 3 & 5 & 7 & 7 & 7 & 15 & 18 & 14 & 14 \\
        \texttt{place\_burger\_fries} & 2 & 14 & 18 & 14 & 17 & 18 & 17 & 19 & 19 & 23 & 23 & 18 & 22 \\
        \texttt{place\_can\_basket} & 1 & 1 & 1 & 2 & 8 & 4 & 12 & 8 & 10 & 14 & 17 & 9 & 17 \\
        \texttt{place\_cans\_plasticbox} & 3 & 6 & 8 & 9 & 5 & 7 & 13 & 17 & 18 & 15 & 20 & 16 & 27 \\
        \texttt{place\_container\_plate} & 0 & 15 & 18 & 18 & 19 & 22 & 23 & 23 & 25 & 28 & 27 & 26 & 26 \\
        \texttt{place\_dual\_shoes} & 1 & 4 & 8 & 3 & 7 & 7 & 11 & 14 & 9 & 14 & 15 & 10 & 20 \\
        \texttt{place\_empty\_cup} & 1 & 3 & 9 & 6 & 9 & 13 & 11 & 14 & 14 & 22 & 21 & 14 & 22 \\
        \texttt{place\_fan} & 0 & 1 & 5 & 3 & 5 & 6 & 3 & 3 & 5 & 9 & 9 & 2 & 4 \\
        \texttt{place\_mouse\_pad} & 0 & 1 & 1 & 0 & 2 & 1 & 2 & 1 & 2 & 3 & 4 & 1 & 4 \\
        \texttt{place\_object\_basket} & 1 & 7 & 10 & 5 & 13 & 13 & 11 & 12 & 16 & 17 & 15 & 9 & 20 \\
        \texttt{place\_object\_scale} & 1 & 2 & 2 & 2 & 4 & 4 & 3 & 2 & 2 & 6 & 6 & 1 & 5 \\
        \texttt{place\_object\_stand} & 1 & 3 & 2 & 6 & 8 & 13 & 7 & 13 & 15 & 14 & 11 & 11 & 14 \\
        \texttt{place\_phone\_stand} & 1 & 0 & 3 & 3 & 3 & 7 & 3 & 8 & 9 & 14 & 13 & 7 & 22 \\
        \texttt{place\_shoe} & 3 & 10 & 12 & 10 & 14 & 7 & 12 & 14 & 13 & 17 & 20 & 7 & 23 \\
        \texttt{press\_stapler} & 6 & 2 & 9 & 7 & 11 & 12 & 13 & 11 & 11 & 17 & 17 & 10 & 19 \\
        \texttt{put\_bottles\_dustbin} & 0 & 0 & 1 & 1 & 1 & 0 & 1 & 1 & 2 & 3 & 2 & 5 & 4 \\
        \texttt{put\_object\_cabinet} & 1 & 9 & 5 & 10 & 10 & 9 & 11 & 10 & 12 & 12 & 11 & 11 & 10 \\
        \texttt{rotate\_qrcode} & 0 & 3 & 5 & 4 & 4 & 7 & 2 & 4 & 5 & 7 & 12 & 2 & 18 \\
        \texttt{scan\_object} & 2 & 4 & 2 & 4 & 4 & 5 & 3 & 4 & 5 & 12 & 7 & 10 & 11 \\
        \texttt{shake\_bottle} & 20 & 25 & 22 & 25 & 28 & 29 & 27 & 30 & 27 & 29 & 30 & 29 & 30 \\
        \texttt{shake\_bottle\_horizontally} & 19 & 24 & 25 & 25 & 28 & 28 & 28 & 30 & 27 & 29 & 30 & 27 & 30 \\
        \texttt{stack\_blocks\_three} & 0 & 1 & 1 & 3 & 2 & 3 & 3 & 3 & 7 & 12 & 17 & 17 & 13 \\
        \texttt{stack\_blocks\_two} & 0 & 5 & 2 & 7 & 11 & 15 & 16 & 19 & 23 & 20 & 20 & 25 & 19 \\
        \texttt{stack\_bowls\_three} & 1 & 3 & 4 & 3 & 7 & 7 & 7 & 13 & 11 & 20 & 16 & 20 & 24 \\
        \texttt{stack\_bowls\_two} & 1 & 6 & 7 & 8 & 15 & 15 & 14 & 19 & 21 & 24 & 25 & 27 & 26 \\
        \texttt{stamp\_seal} & 0 & 0 & 0 & 2 & 1 & 2 & 1 & 1 & 3 & 11 & 4 & 5 & 9 \\
        \texttt{turn\_switch} & 1 & 3 & 8 & 8 & 9 & 9 & 6 & 7 & 5 & 8 & 13 & 6 & 11 \\
        \midrule
        \textbf{Mean (\%)}
            & 7.6 & 17.0 & 20.6 & 22.2 & 27.4
            & 29.5 & 30.1 & 34.8 & 36.9
            & 50.6 & 53.4 & 41.2 & 58.1 \\
        \bottomrule
    \end{tabular}
\end{table*}

\begin{table*}[p]
    \centering
    \footnotesize
    \setlength{\tabcolsep}{2.8pt}
    \renewcommand{\arraystretch}{0.97}
    \caption{
        \textbf{Complete RoboTwin clean results.}
        Successful trials out of 30 for each task and
        training condition.
        The final row gives aggregate success rates (\%).
    }
    \label{tab:app_robotwin_clean}
    \begin{tabular}{@{}l*{13}{c}@{}}
        \toprule
        & & \multicolumn{2}{c}{RQ1}
          & \multicolumn{8}{c}{RQ2}
          & \multicolumn{2}{c}{RQ3} \\
        \cmidrule(lr){3-4}
        \cmidrule(lr){5-12}
        \cmidrule(l){13-14}
        Task & Std & Lat & MM & MAE & RE & Ber
             & M-SC & B-SC & Mix & S-SC & S-S & S-M & S-A \\
        \midrule
        \texttt{adjust\_bottle} & 30 & 30 & 30 & 30 & 30 & 30 & 30 & 30 & 30 & 30 & 30 & 30 & 30 \\
        \texttt{beat\_block\_hammer} & 28 & 26 & 29 & 29 & 25 & 29 & 30 & 28 & 29 & 28 & 27 & 28 & 27 \\
        \texttt{blocks\_ranking\_rgb} & 26 & 28 & 28 & 29 & 28 & 29 & 28 & 29 & 30 & 29 & 29 & 28 & 29 \\
        \texttt{blocks\_ranking\_size} & 11 & 15 & 18 & 19 & 21 & 16 & 14 & 21 & 20 & 18 & 14 & 21 & 15 \\
        \texttt{click\_alarmclock} & 30 & 30 & 30 & 30 & 30 & 30 & 30 & 30 & 30 & 30 & 30 & 30 & 30 \\
        \texttt{click\_bell} & 30 & 30 & 30 & 30 & 30 & 30 & 30 & 30 & 30 & 30 & 30 & 30 & 30 \\
        \texttt{dump\_bin\_bigbin} & 23 & 26 & 28 & 25 & 24 & 26 & 26 & 27 & 27 & 29 & 28 & 27 & 27 \\
        \texttt{grab\_roller} & 30 & 30 & 30 & 30 & 30 & 30 & 30 & 30 & 30 & 30 & 30 & 30 & 30 \\
        \texttt{handover\_block} & 11 & 16 & 21 & 22 & 19 & 20 & 12 & 23 & 19 & 17 & 17 & 24 & 21 \\
        \texttt{handover\_mic} & 16 & 27 & 29 & 29 & 29 & 30 & 29 & 30 & 30 & 30 & 30 & 29 & 30 \\
        \texttt{hanging\_mug} & 6 & 7 & 8 & 5 & 7 & 5 & 5 & 7 & 10 & 7 & 5 & 10 & 9 \\
        \texttt{lift\_pot} & 26 & 26 & 27 & 27 & 25 & 25 & 25 & 25 & 25 & 26 & 26 & 28 & 27 \\
        \texttt{move\_can\_pot} & 28 & 26 & 26 & 26 & 27 & 27 & 23 & 27 & 26 & 24 & 25 & 25 & 25 \\
        \texttt{move\_pillbottle\_pad} & 28 & 28 & 28 & 29 & 28 & 29 & 30 & 30 & 28 & 29 & 30 & 25 & 26 \\
        \texttt{move\_playingcard\_away} & 27 & 29 & 28 & 28 & 30 & 28 & 28 & 29 & 30 & 30 & 28 & 30 & 30 \\
        \texttt{move\_stapler\_pad} & 7 & 5 & 7 & 9 & 4 & 12 & 9 & 8 & 9 & 13 & 8 & 14 & 9 \\
        \texttt{open\_laptop} & 29 & 29 & 29 & 26 & 30 & 30 & 30 & 27 & 30 & 28 & 30 & 28 & 29 \\
        \texttt{open\_microwave} & 15 & 5 & 17 & 16 & 13 & 17 & 23 & 19 & 20 & 23 & 23 & 10 & 24 \\
        \texttt{pick\_diverse\_bottles} & 9 & 4 & 9 & 10 & 8 & 9 & 9 & 11 & 7 & 7 & 7 & 10 & 8 \\
        \texttt{pick\_dual\_bottles} & 18 & 12 & 17 & 17 & 16 & 18 & 15 & 19 & 12 & 18 & 12 & 20 & 20 \\
        \texttt{place\_a2b\_left} & 24 & 25 & 25 & 22 & 28 & 23 & 22 & 23 & 25 & 23 & 23 & 23 & 23 \\
        \texttt{place\_a2b\_right} & 21 & 24 & 24 & 27 & 24 & 22 & 23 & 23 & 23 & 23 & 24 & 23 & 22 \\
        \texttt{place\_bread\_basket} & 23 & 26 & 25 & 27 & 26 & 25 & 28 & 27 & 26 & 26 & 29 & 26 & 25 \\
        \texttt{place\_bread\_skillet} & 20 & 25 & 24 & 26 & 24 & 28 & 25 & 28 & 27 & 25 & 27 & 28 & 28 \\
        \texttt{place\_burger\_fries} & 30 & 30 & 30 & 28 & 29 & 30 & 26 & 28 & 29 & 29 & 29 & 29 & 27 \\
        \texttt{place\_can\_basket} & 25 & 21 & 22 & 17 & 19 & 20 & 24 & 26 & 17 & 20 & 22 & 22 & 24 \\
        \texttt{place\_cans\_plasticbox} & 30 & 30 & 29 & 30 & 30 & 29 & 30 & 28 & 30 & 30 & 30 & 30 & 30 \\
        \texttt{place\_container\_plate} & 30 & 29 & 30 & 29 & 29 & 30 & 30 & 30 & 30 & 30 & 30 & 30 & 30 \\
        \texttt{place\_dual\_shoes} & 21 & 25 & 21 & 24 & 22 & 23 & 22 & 23 & 24 & 23 & 25 & 22 & 26 \\
        \texttt{place\_empty\_cup} & 29 & 29 & 29 & 29 & 29 & 29 & 29 & 29 & 29 & 29 & 29 & 29 & 30 \\
        \texttt{place\_fan} & 19 & 13 & 24 & 22 & 22 & 26 & 17 & 24 & 22 & 21 & 20 & 21 & 23 \\
        \texttt{place\_mouse\_pad} & 16 & 13 & 15 & 17 & 12 & 12 & 16 & 17 & 14 & 15 & 17 & 20 & 10 \\
        \texttt{place\_object\_basket} & 16 & 28 & 24 & 22 & 27 & 24 & 26 & 24 & 24 & 25 & 23 & 22 & 26 \\
        \texttt{place\_object\_scale} & 23 & 20 & 20 & 19 & 22 & 23 & 23 & 25 & 23 & 22 & 22 & 23 & 22 \\
        \texttt{place\_object\_stand} & 28 & 27 & 27 & 26 & 25 & 27 & 25 & 25 & 29 & 28 & 27 & 29 & 27 \\
        \texttt{place\_phone\_stand} & 27 & 28 & 28 & 27 & 28 & 29 & 28 & 29 & 27 & 27 & 27 & 28 & 26 \\
        \texttt{place\_shoe} & 24 & 24 & 26 & 29 & 24 & 26 & 25 & 26 & 25 & 23 & 24 & 26 & 25 \\
        \texttt{press\_stapler} & 27 & 25 & 27 & 20 & 26 & 24 & 24 & 29 & 25 & 22 & 24 & 24 & 24 \\
        \texttt{put\_bottles\_dustbin} & 3 & 7 & 17 & 15 & 16 & 20 & 19 & 20 & 18 & 20 & 20 & 21 & 16 \\
        \texttt{put\_object\_cabinet} & 13 & 15 & 15 & 13 & 12 & 15 & 16 & 16 & 14 & 16 & 15 & 11 & 14 \\
        \texttt{rotate\_qrcode} & 21 & 22 & 20 & 24 & 24 & 26 & 22 & 23 & 22 & 24 & 19 & 24 & 23 \\
        \texttt{scan\_object} & 22 & 24 & 22 & 19 & 20 & 21 & 20 & 22 & 21 & 19 & 20 & 19 & 20 \\
        \texttt{shake\_bottle} & 30 & 30 & 30 & 30 & 30 & 30 & 30 & 30 & 30 & 30 & 30 & 30 & 30 \\
        \texttt{shake\_bottle\_horizontally} & 30 & 30 & 30 & 30 & 30 & 30 & 30 & 30 & 30 & 30 & 30 & 30 & 30 \\
        \texttt{stack\_blocks\_three} & 27 & 29 & 27 & 29 & 25 & 26 & 26 & 28 & 25 & 28 & 25 & 26 & 21 \\
        \texttt{stack\_blocks\_two} & 30 & 29 & 30 & 30 & 30 & 29 & 30 & 30 & 29 & 30 & 29 & 30 & 27 \\
        \texttt{stack\_bowls\_three} & 24 & 24 & 26 & 24 & 26 & 21 & 24 & 24 & 23 & 25 & 21 & 24 & 25 \\
        \texttt{stack\_bowls\_two} & 28 & 28 & 29 & 28 & 28 & 30 & 30 & 29 & 29 & 29 & 30 & 26 & 27 \\
        \texttt{stamp\_seal} & 21 & 21 & 21 & 22 & 18 & 19 & 22 & 25 & 17 & 22 & 17 & 22 & 24 \\
        \texttt{turn\_switch} & 13 & 15 & 18 & 17 & 17 & 17 & 23 & 19 & 16 & 19 & 21 & 18 & 22 \\
        \midrule
        \textbf{Mean (\%)}
            & 74.9 & 76.3 & 80.3 & 79.2
            & 78.4 & 80.3 & 79.4 & 82.7
            & 79.7 & 80.6 & 79.2 & 80.9 & 80.2 \\
        \bottomrule
    \end{tabular}
\end{table*}

% ============================================================
% C.2 RoboCasa
% ============================================================

\subsection{RoboCasa Task-Level Results}
\label{app:robocasa_results}

Table~\ref{tab:app_robocasa_tasks} reports
success counts for all 18 RoboCasa tasks,
with 10 trials per task and condition.
Unlike RoboTwin BG OOD, RoboCasa uses
randomized-background evaluation without
a strict held-out appearance split.

For RQ1, MAE-Matched outperforms latent
masking in aggregate (59.4\% vs.\ 51.1\%),
improving nine tasks, tying four, and
performing worse on five.

For RQ2, shared-color MAE and Bernoulli
masking outperform their independent-color
counterparts on 9 and 11 tasks, respectively.
SDO-SC achieves 70.6\% success, compared
with 68.3\% for textured SDO, with 7/4/7
improved/tied/worse tasks.
Thus, removing natural-object textures
preserves comparable aggregate performance.
Unlike RoboTwin, the advantage of SDO-SC
over Bernoulli-SC is small
(70.6\% vs.\ 68.9\%), while Mixed Masking
falls below Bernoulli-SC (62.2\% vs.\ 68.9\%).

For RQ3, single-view SDO exceeds
matched-budget masking in aggregate
(68.3\% vs.\ 66.7\%), with 5/7/6
improved/tied/worse tasks.
Full-strength all-view SDO achieves the
highest aggregate success (73.3\%),
outperforming single-view SDO on eight
tasks, tying on four, and performing
worse on six.
These comparisons preserve the aggregate
ordering observed on RoboTwin, but with
smaller differences between allocation
strategies.

\begin{table*}[p]
    \centering
    \footnotesize
    \setlength{\tabcolsep}{2.4pt}
    \renewcommand{\arraystretch}{0.97}
    \caption{
        \textbf{Complete RoboCasa results.}
        Successful trials out of 10 for each of the
        18 tasks under randomized-background evaluation.
    }
    \label{tab:app_robocasa_tasks}
    \begin{tabular}{@{}l*{13}{c}@{}}
        \toprule
        & & \multicolumn{2}{c}{RQ1}
          & \multicolumn{8}{c}{RQ2}
          & \multicolumn{2}{c}{RQ3} \\
        \cmidrule(lr){3-4}
        \cmidrule(lr){5-12}
        \cmidrule(l){13-14}
        Task & Std & Lat & MM & MAE & RE & Ber
             & M-SC & B-SC & Mix & S-SC & S-S & S-M & S-A \\
        \midrule
        \texttt{CloseBlenderLid}
            & 0 & 0 & 2 & 1 & 0 & 0 & 1 & 0 & 0
            & 1 & 0 & 0 & 1 \\
        \texttt{CloseFridge}
            & 4 & 5 & 4 & 6 & 6 & 7 & 5 & 8 & 5
            & 4 & 5 & 6 & 9 \\
        \texttt{CloseToasterOvenDoor}
            & 5 & 5 & 4 & 6 & 6 & 8 & 7 & 4 & 6
            & 9 & 6 & 4 & 6 \\
        \texttt{CoffeeSetupMug}
            & 8 & 5 & 8 & 6 & 8 & 6 & 9 & 9 & 6
            & 9 & 6 & 6 & 5 \\
        \texttt{NavigateKitchen}
            & 0 & 0 & 0 & 0 & 0 & 0 & 1 & 1 & 1
            & 1 & 4 & 2 & 6 \\
        \texttt{OpenCabinet}
            & 8 & 6 & 8 & 8 & 5 & 8 & 10 & 9 & 9
            & 9 & 10 & 10 & 9 \\
        \texttt{OpenDrawer}
            & 8 & 7 & 8 & 9 & 5 & 7 & 10 & 10 & 9
            & 9 & 10 & 10 & 10 \\
        \texttt{OpenStandMixerHead}
            & 9 & 8 & 10 & 10 & 10 & 10 & 10 & 10 & 10
            & 10 & 10 & 10 & 10 \\
        \texttt{PickPlaceCounterToCabinet}
            & 6 & 6 & 6 & 5 & 6 & 8 & 6 & 7 & 4
            & 7 & 8 & 6 & 6 \\
        \texttt{PickPlaceCounterToStove}
            & 8 & 9 & 8 & 8 & 8 & 8 & 9 & 9 & 9
            & 9 & 10 & 7 & 8 \\
        \texttt{PickPlaceDrawerToCounter}
            & 3 & 4 & 7 & 5 & 7 & 5 & 8 & 6 & 6
            & 8 & 6 & 8 & 8 \\
        \texttt{PickPlaceSinkToCounter}
            & 6 & 4 & 6 & 5 & 4 & 7 & 6 & 8 & 9
            & 6 & 5 & 7 & 7 \\
        \texttt{PickPlaceToasterToCounter}
            & 5 & 1 & 8 & 8 & 8 & 8 & 7 & 10 & 8
            & 10 & 9 & 10 & 9 \\
        \texttt{SlideDishwasherRack}
            & 8 & 7 & 6 & 8 & 7 & 6 & 8 & 7 & 7
            & 7 & 9 & 7 & 8 \\
        \texttt{TurnOffStove}
            & 2 & 2 & 3 & 3 & 3 & 4 & 2 & 3 & 3
            & 4 & 4 & 4 & 5 \\
        \texttt{TurnOnElectricKettle}
            & 8 & 7 & 7 & 8 & 9 & 9 & 8 & 9 & 8
            & 9 & 9 & 9 & 7 \\
        \texttt{TurnOnMicrowave}
            & 3 & 7 & 3 & 6 & 6 & 5 & 6 & 5 & 4
            & 7 & 4 & 5 & 9 \\
        \texttt{TurnOnSinkFaucet}
            & 6 & 9 & 9 & 8 & 8 & 8 & 7 & 9 & 8
            & 8 & 8 & 9 & 9 \\
        \midrule
        \textbf{Mean (\%)}
            & 53.9 & 51.1 & 59.4 & 61.1
            & 58.9 & 63.3 & 66.7 & 68.9
            & 62.2 & 70.6 & 68.3 & 66.7 & 73.3 \\
        \bottomrule
    \end{tabular}
\end{table*}

% ============================================================
% C.3 Paired Task-Level Comparisons
% ============================================================

\subsection{Paired Task-Level Comparisons}
\label{app:statistics}

Table~\ref{tab:app_task_comparisons} summarizes
paired success-rate differences averaged across
tasks. I/T/W counts improved, tied, and worse
tasks. Descriptive 95\% percentile intervals
use 20,000 paired task-bootstrap resamples
(seed 2027), capturing task-level variation
rather than training-seed uncertainty.

For RQ1, MAE-Matched exceeds latent masking
by 3.6 pp on RoboTwin (95\% CI: [1.5, 5.7])
and 8.3 pp on RoboCasa ([-1.7, 18.9]),
with greater task-level consistency on RoboTwin.

For RQ2, SDO-SC exceeds Bernoulli-SC by
15.8 pp on RoboTwin ([12.1, 19.7]) but only
1.7 pp on RoboCasa ([-6.1, 10.0]).
Shared-color variants outperform independent
coloring in aggregate on both benchmarks.
Removing natural-object textures changes
success by -2.8 and +2.2 pp, respectively,
with intervals spanning zero.
Mixed Masking exceeds Bernoulli-SC on RoboTwin
but underperforms it on RoboCasa.

For RQ3, single-view SDO exceeds matched-budget
allocation by 12.2 pp on RoboTwin ([7.3, 17.2])
and 1.7 pp on RoboCasa ([-4.4, 8.3]).
Full-strength all-view SDO achieves the
highest aggregate success on both benchmarks,
exceeding single-view SDO by 4.7 pp
([1.4, 7.9]) and 5.0 pp ([-3.3, 13.9]),
respectively.

\begin{table*}[t]
    \centering
    \footnotesize
    \setlength{\tabcolsep}{7pt}
    \renewcommand{\arraystretch}{1.06}
    \caption{
        \textbf{Paired task-level method comparisons.}
        $\Delta$ is the mean success-rate difference
        (first minus second method) in percentage points.
        I/T/W denotes numbers of improved, tied, and
        worse tasks.
        CIs use 20,000 percentile task-bootstrap
        resamples and do not measure
        training-seed uncertainty.
    }
    \label{tab:app_task_comparisons}
    \begin{tabular}{@{}llrrl@{}}
        \toprule
        Axis & Comparison & $\Delta$ (pp)
             & I/T/W & 95\% CI (pp) \\
        \midrule
        \multicolumn{5}{@{}l}{
            \textit{RoboTwin BG OOD (50 tasks)}
        } \\
        RQ1 & MM $-$ Lat & +3.6 & 27/14/9
            & [1.5, 5.7] \\
        RQ2 & M-SC $-$ MAE & +7.9 & 34/7/9
            & [5.1, 10.9] \\
        RQ2 & B-SC $-$ Ber & +5.3 & 30/5/15
            & [2.3, 8.5] \\
        RQ2 & Mix $-$ B-SC & +2.1 & 28/7/15
            & [-0.7, 5.0] \\
        RQ2 & S-SC $-$ B-SC
    & +15.8 & 42/3/5 & [12.1, 19.7] \\
        RQ2 & S-SC $-$ S-S & -2.8 & 16/7/27
            & [-6.0, 0.5] \\
        RQ3 & S-S $-$ S-M & +12.2 & 34/4/12
            & [7.3, 17.2] \\
        RQ3 & S-A $-$ S-S & +4.7 & 27/7/16
            & [1.4, 7.9] \\
        RQ3 & S-A $-$ S-M & +16.9 & 38/3/9
            & [11.5, 22.5] \\
        Overall & S-A $-$ Std & +50.5 & 50/0/0
            & [43.7, 57.1] \\
        \midrule
        \multicolumn{5}{@{}l}{
            \textit{RoboCasa (18 tasks)}
        } \\
        RQ1 & MM $-$ Lat & +8.3 & 9/4/5
            & [-1.7, 18.9] \\
        RQ2 & M-SC $-$ MAE & +5.6 & 9/5/4
            & [0.0, 11.1] \\
        RQ2 & B-SC $-$ Ber & +5.6 & 11/4/3
            & [-1.7, 12.2] \\
        RQ2 & Mix $-$ B-SC & -6.7 & 2/8/8
            & [-12.8, -0.6] \\
        RQ2 & S-SC $-$ B-SC
    & +1.7 & 5/9/4 & [-6.1, 10.0] \\
        RQ2 & S-SC $-$ S-S & +2.2 & 7/4/7
            & [-5.6, 10.0] \\
        RQ3 & S-S $-$ S-M & +1.7 & 5/7/6
            & [-4.4, 8.3] \\
        RQ3 & S-A $-$ S-S & +5.0 & 8/4/6
            & [-3.3, 13.9] \\
        RQ3 & S-A $-$ S-M & +6.7 & 8/6/4
            & [-0.6, 14.4] \\
        Overall & S-A $-$ Std & +19.4 & 13/3/2
            & [8.3, 30.6] \\
        \bottomrule
    \end{tabular}
\end{table*}

\subsection{Robustness to Additional Visual Shifts}
\label{app:additional_shifts}

We further evaluate all-view SDO under lighting
and clutter shifts on RoboTwin, beyond the
background changes considered in the main study.
Compared with Standard, all-view SDO improves
success under lighting shifts from 65.9\%
to 77.5\% (+11.6 pp) and under clutter shifts
from 18.6\% to 62.5\% (+43.9 pp).
These results show that the robustness gains
extend beyond background appearance changes
to lighting variations and additional
task-irrelevant objects.

\subsection{Background Sensitivity Analysis}
\label{app:bg_sensitivity_detail}

Table~\ref{tab:app_bg_sensitivity} reports
background-induced action sensitivity
and BG OOD success for all 13 RoboTwin training
configurations. The 10-task success rates
are computed on the same tasks used to
measure $D_{\mathrm{bg}}$.

The strong negative association persists
on this matched subset (Pearson
$r=-0.974$, Spearman $\rho=-0.956$).
It also holds within the eight RQ2
configurations ($r=-0.933$,
$\rho=-0.810$). These results show that
the relationship extends beyond the
full-benchmark averages and remains
visible among within-view corruption
designs.

\paragraph{Leave-one-configuration-out analysis.}
We further assess the consistency of this
relationship using linear regression with
$D_{\mathrm{bg}}$ as the sole predictor.
Each fit uses 12 configurations to predict
BG OOD success for the held-out configuration.
The mean absolute error is 3.8 pp for
50-task success and 4.3 pp for the matched
10-task subset. These results characterize
predictive consistency among the evaluated
configurations, rather than performance
on unseen corruption families.

\clearpage
\onecolumn

% ============================================================
% Appendix D: Real-Robot Experiments
% ============================================================
\begin{figure*}[!htbp]
    \centering
    \setlength{\tabcolsep}{1pt}
    \renewcommand{\arraystretch}{1.0}

    \begin{tabular}{@{}cccc@{}}

        \multicolumn{4}{@{}l@{}}{
            \textbf{Blue BG}
            \quad Red Block to Plate
        } \\[2pt]

        \includegraphics[width=0.245\textwidth]{
            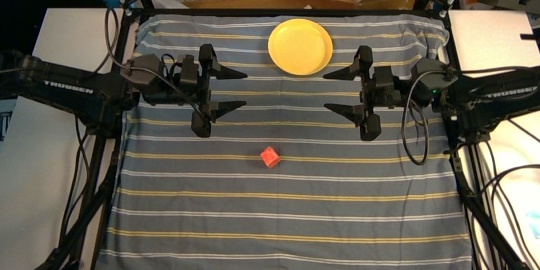
        } &
        \includegraphics[width=0.245\textwidth]{
            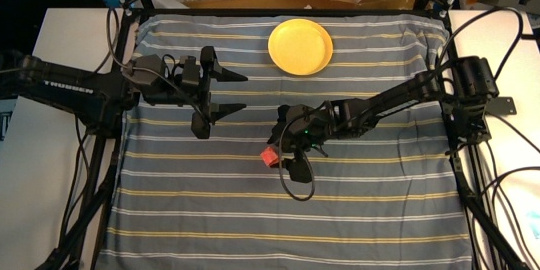
        } &
        \includegraphics[width=0.245\textwidth]{
            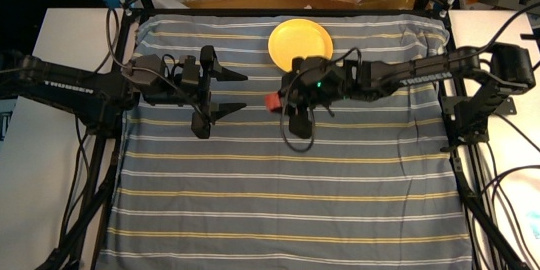
        } &
        \includegraphics[width=0.245\textwidth]{
            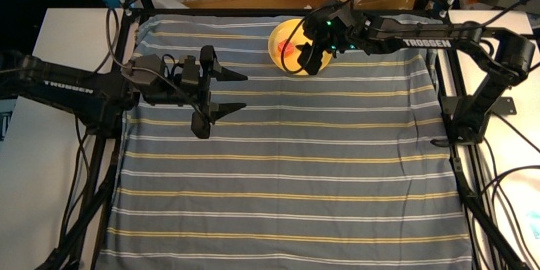
        } \\[5pt]

        \multicolumn{4}{@{}l@{}}{
            \textbf{Brown BG}
            \quad Cake Handover to Plate
        } \\[2pt]

        \includegraphics[width=0.245\textwidth]{
            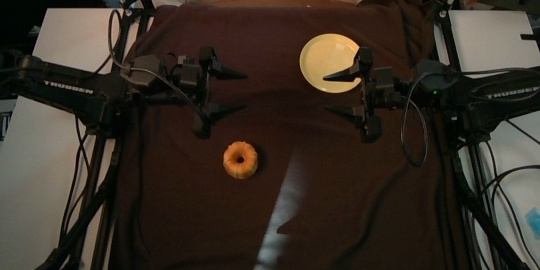
        } &
        \includegraphics[width=0.245\textwidth]{
            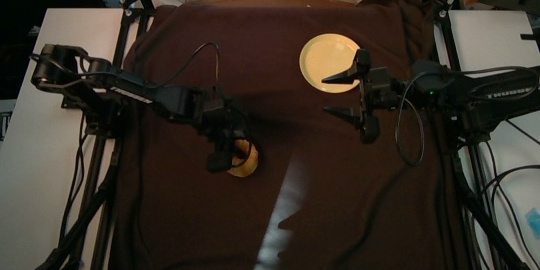
        } &
        \includegraphics[width=0.245\textwidth]{
            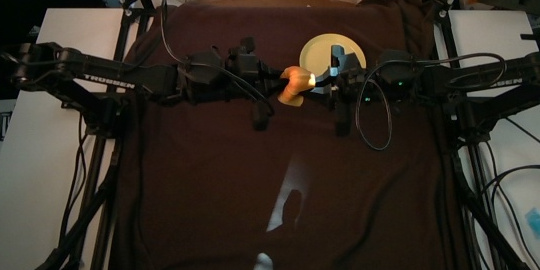
        } &
        \includegraphics[width=0.245\textwidth]{
            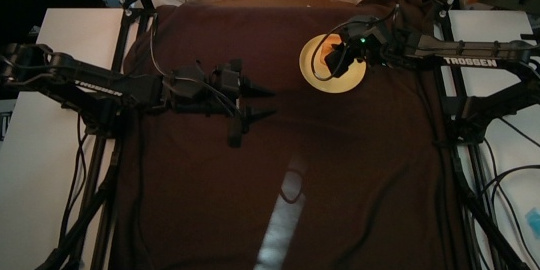
        } \\[5pt]

        \multicolumn{4}{@{}l@{}}{
            \textbf{Low Light}
            \quad Sequential Pick to Plate
        } \\[2pt]

        \includegraphics[width=0.245\textwidth]{
            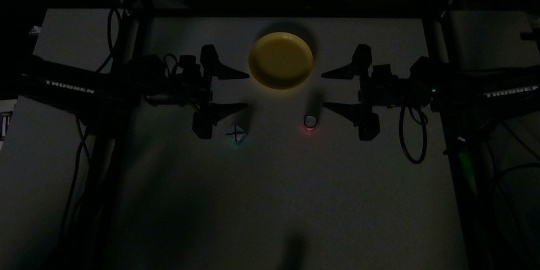
        } &
        \includegraphics[width=0.245\textwidth]{
            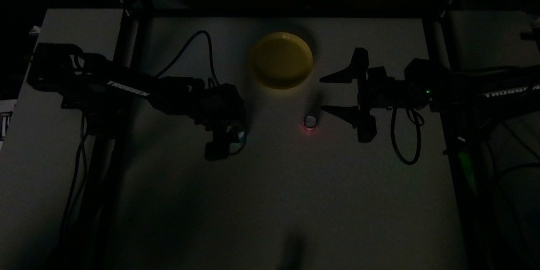
        } &
        \includegraphics[width=0.245\textwidth]{
            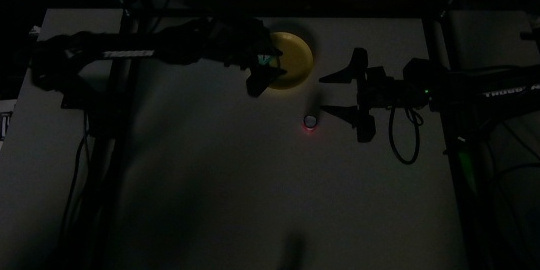
        } &
        \includegraphics[width=0.245\textwidth]{
            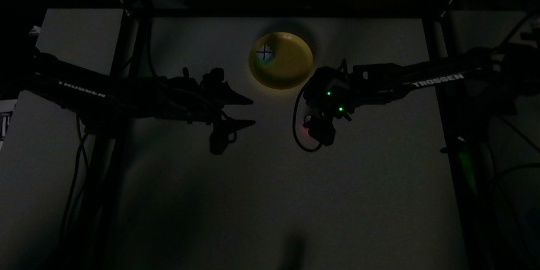
        } \\[5pt]

        \multicolumn{4}{@{}l@{}}{
            \textbf{Distractors}
            \quad Lift Black Bag
        } \\[2pt]

        \includegraphics[width=0.245\textwidth]{
            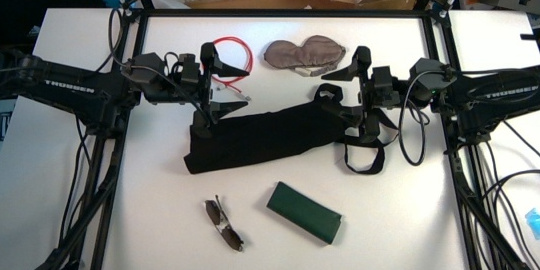
        } &
        \includegraphics[width=0.245\textwidth]{
            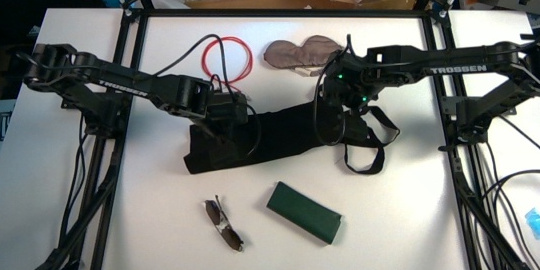
        } &
        \includegraphics[width=0.245\textwidth]{
            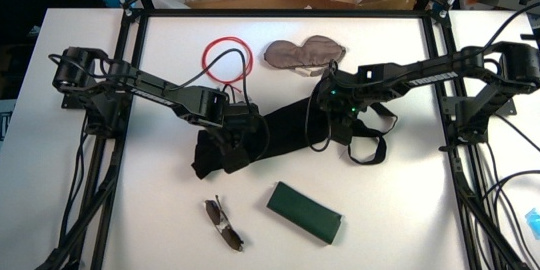
        } &
        \includegraphics[width=0.245\textwidth]{
            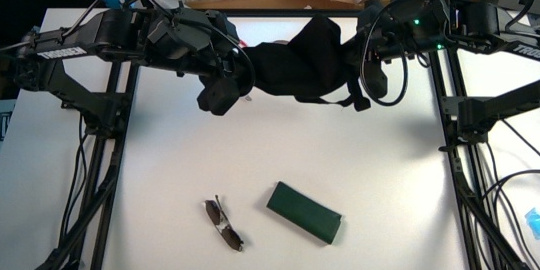
        } \\

    \end{tabular}

    \caption{
        \textbf{Representative real-robot executions
        under four distribution shifts.}
        Each row shows four chronologically ordered
        frames from a successful rollout of the
        occlusion-trained policy.
        From top to bottom:
        Red Block to Plate under Blue BG,
        Cake Handover to Plate under Brown BG,
        Sequential Pick to Plate under Low Light,
        and Lift Black Bag under Distractors.
        Test-time observations remain unmasked.
        Complete success rates, including unsuccessful
        trials, are reported in
        Table~\ref{tab:real_robot_full}.
    }
    \label{fig:app_real_robot_ood_rollouts}
\end{figure*}

% \clearpage

\section{Real-Robot Experimental Details}
\label{app:real_robot}
We collect 50 demonstrations per task
(250 in total) on the TROSSEN platform.
Both Standard and occlusion-trained policies
are trained for 15,000 optimization steps
using the same demonstrations and architecture.

We evaluate each policy under the original
training appearance and four visual shifts:
Blue BG, Brown BG, Low Light, and Distractors.
Each task is evaluated for 30 trials per
policy and condition, resulting in 1,500
physical trials. Both policies share the
same inference procedure, and no artificial
occlusion is applied during evaluation.
Complete task-level results are reported
in Tab.~\ref{tab:real_robot_full}.
% We collect 50 demonstrations for each tasks on the
% TROSSEN platform and evaluate under the original training appearance
% and four distribution shifts: Blue BG, Brown BG,
% Low Light, and Distractors. Each task is evaluated
% for 30 trials per policy and condition, resulting
% in 1,500 physical trials across the standard and
% occlusion-trained policies.

% Table~\ref{tab:real_robot_full} provides the complete
% task-level success rates. The two policies use the
% same demonstrations, architecture, and inference
% procedure, differing only in training-time visual
% corruption. For training, we train 15k steps. No artificial occlusion is applied
% during evaluation.

\paragraph{Qualitative execution examples.}
Figure~\ref{fig:app_real_robot_ood_rollouts} shows
representative successful executions of the
occlusion-trained policy under the four OOD conditions.
Each row presents four temporally ordered frames
from a single rollout. The examples cover object
placement, bimanual handover, sequential manipulation,
and cooperative lifting. These qualitative examples
complement the complete task-level results in
Table~\ref{tab:real_robot_full}.

\begin{table}[t]
    \centering
    \caption{
        \textbf{Real-robot manipulation success rates (\%).}
        We evaluate five tasks under the original
        training appearance (Clean) and four
        distribution shifts.
        Each task--method--condition combination
        contains 30 trials.
        OOD Avg. averages the four shifted conditions.
        $\Delta$ denotes the improvement from
        occlusion training in percentage points.
    }
    \label{tab:real_robot_full}

    \footnotesize
    \setlength{\tabcolsep}{4pt}

 \begin{tabular*}{\textwidth}{
    @{\extracolsep{\fill}}llcccccc@{}
}
        \toprule
        Task
        & Method
        & Clean
        & Blue BG
        & Brown BG
        & Low Light
        & Distractors
        & OOD Avg. \\
        \midrule

        Pick Red-Circle Cube
        & Standard
        & 46.7 & 0.0 & 0.0 & 3.3 & 13.3 & 4.2 \\
        & Occlusion
        & 56.7 & 46.7 & 30.0 & 40.0 & 26.7 & 35.8 \\

        \addlinespace[3pt]

        Red Block to Plate
        & Standard
        & 90.0 & 3.3 & 0.0 & 6.7 & 3.3 & 3.3 \\
        & Occlusion
        & 96.7 & 73.3 & 40.0 & 60.0 & 26.7 & 50.0 \\

        \addlinespace[3pt]

        Lift Black Bag
        & Standard
        & 93.3 & 0.0 & 3.3 & 16.7 & 90.0 & 27.5 \\
        & Occlusion
        & 96.7 & 93.3 & 90.0 & 93.3 & 90.0 & 91.7 \\

        \addlinespace[3pt]

        Cake Handover
        & Standard
        & 86.7 & 0.0 & 3.3 & 10.0 & 73.3 & 21.7 \\
        & Occlusion
        & 90.0 & 90.0 & 86.7 & 83.3 & 80.0 & 85.0 \\

        \addlinespace[3pt]

        Sequential Pick
        & Standard
        & 86.7 & 0.0 & 0.0 & 13.3 & 66.7 & 20.0 \\
        & Occlusion
        & 90.0 & 86.7 & 80.0 & 86.7 & 56.7 & 77.5 \\

        \midrule

        \textbf{Average}
        & Standard
        & 80.7 & 0.7 & 1.3 & 10.0 & 49.3 & 15.3 \\

        & \textbf{Occlusion}
        & \textbf{86.0}
        & \textbf{78.0}
        & \textbf{65.3}
        & \textbf{72.7}
        & \textbf{56.0}
        & \textbf{68.0} \\

        \addlinespace[2pt]

        & $\boldsymbol{\Delta}$
        & \textbf{+5.3}
        & \textbf{+77.3}
        & \textbf{+64.0}
        & \textbf{+62.7}
        & \textbf{+6.7}
        & \textbf{+52.7} \\

        \bottomrule
    \end{tabular*}
\end{table}

\end{document}